\documentclass[12pt]{elsarticle}
\usepackage[utf8]{inputenc}
\usepackage{times}
\usepackage{hyperref}       
\usepackage{url}            
\usepackage{color}
\usepackage{graphicx}
\usepackage[left=2.5cm,right=2.5cm,top=2.5cm,height=24cm]{geometry}
\usepackage{bm}

\graphicspath{ {FIGS/} {FIGS/benthamSamples/} {FIGS/EvalExample/} }
\hypersetup{
  pdftitle={End-to-End Page-Level Assessment of HTR},
  pdfauthor={Vidal, Toselli, Ríos, Calvo},
  pdfsubject={Evaluation metrics},
  pdfcreator={PRHLT Research Center},
  pdfkeywords={Evaluation, Handwritten Text Recognition},
  pdfpagemode={UseNone},
  bookmarks,
  bookmarksopen,
  pdfstartview={FitV},
  colorlinks,
  linkcolor={blue},
  citecolor={red},
  urlcolor=[rgb]{0,0.4,0.4} 
}

\usepackage{array}
\usepackage{multirow}
\usepackage{booktabs}

\definecolor{grey}{rgb}{0.5,0.5,0.5}
\definecolor{darkred}{rgb}{0.6,0.1,0.1}

\newcommand{\red}[1]{\textcolor{red}{#1}}
\newcommand{\blue}[1]{\textcolor{blue}{#1}}

\usepackage{amsmath}
\DeclareMathOperator*{\WER}{\mathrm{WER}}
\DeclareMathOperator*{\tWER}{\mathrm{tWER}}
\DeclareMathOperator*{\CER}{\mathrm{CER}}
\DeclareMathOperator*{\tCER}{\mathrm{tCER}}
\DeclareMathOperator*{\hCER}{\mathrm{hCER}}
\DeclareMathOperator*{\bWER}{\mathrm{bWER}}
\DeclareMathOperator*{\hWER}{\mathrm{hWER}}

\DeclareMathOperator*{\eqdef}{\stackrel{\text{\tiny def}}{=}}

\DeclareMathOperator*{\edit}{\mathrm{d}}

\usepackage{upgreek}

\title{~\\[-1.5em]
  End-to-End Page-Level Assessment of\\Handwritten Text Recognition\\[-0.3em]}
\author{%
  Enrique Vidal,$\!^{1,2}$ Alejandro H. Toselli,$\!^1$
  Antonio R\'{i}os-Vila,$\!^3$ Jorge Calvo-Zaragoza$^3$\\[0.1em]
  {\normalsize $^1$PRHLT Research Center, Universitat Politècnica de València}\\
  {\normalsize $^2$Valencian Graduate School and Research Network of
                   Artificial Intelligence (valgrAI)}\\
  {\normalsize $^3$U.I. for Computer Research, University of Alicante}\\[-1em]
\vspace{-2.5em}
}
\date{}
\begin{document}
\begin{frontmatter}
\begin{abstract}
The evaluation of Handwritten Text Recognition (HTR) systems has
traditionally used metrics based on the edit distance between HTR and
ground truth (GT) transcripts, at both the character and word
levels. This is very adequate when the experimental protocol assumes
that both GT and HTR text lines are the same, which allows edit
distances to be independently computed to each given line.
%
Driven by recent advances in pattern recognition, HTR systems
increasingly face the end-to-end page-level transcription of a
document, where the precision of locating the different text lines and
their corresponding reading order (RO) play a key role. In such a
case, the standard metrics do not take into account the
inconsistencies that might appear.  In this paper, the problem of
evaluating HTR systems at the page level is introduced in detail.  We
analyse the convenience of using a two-fold evaluation, where the
transcription accuracy and the RO goodness are considered separately.
Different alternatives are proposed, analysed and empirically
compared both through partially simulated and through real, full
end-to-end experiments.
Results support the validity of the proposed two-fold evaluation
approach.
An important conclusion is that such an evaluation can be adequately
achieved by just two simple and well-known metrics: the Word Error
Rate (WER), that takes transcription sequentiality into account, and
the here re-formulated Bag of Words Word Error Rate (bWER), that
ignores order.
While the latter directly and very accurately assess intrinsic word
recognition errors, the difference between both metrics ($\Delta$WER)
gracefully correlates with the Normalised Spearman's Foot Rule
Distance (NSFD), a metric which explicitly measures RO errors
associated with layout analysis flaws.
To arrive to these conclusions, we have introduced another metric
called Hungarian Word Word Rate (hWER), based on a here proposed
regularised version of the Hungarian Algorithm.  This metric is shown
to be always almost identical to bWER and both bWER and hWER are also
almost identical to WER whenever HTR transcripts and GT references are
guarantee to be in the same RO.
%
\\[0.5em]
\textbf{Keywords:}
Handwritten Text Recognition; Full-Page End-to-End Text Image
Transcription; Layout Analysis; Text Line Detection; Reading Order;
Evaluation Measures; Bag of Words; Regularised Hungarian Algorithm;
Word Error Rate.

\end{abstract}
\end{frontmatter}


\section{Introduction}\label{sec:intro}

Archives and libraries throughout the world hold billions of
historical manuscripts.  Many of these documents are already digitised
into images, but their access is limited because the contents are not
available in a symbolic format that would allow modern treatment of
textual matters such as editing, indexing, and retrieval. Handwritten
Text Recognition (HTR)%
\footnote{While all the problems and methods discussed in this paper
  equally apply to \emph{printed} text and OCR transcripts, we keep
  the main focus on \emph{handwritten} text, where the problems become
  more insidious and the solutions more relevant.
} %
is the cornerstone in this situation which aims to provide automatic
ways of transcribing these
documents~\cite{muehlberger2019transforming}.

In classical HTR laboratory experiments, the text lines are assumed
to be given. Therefore, the performance is evaluated at the line
level. Traditional evaluation measures for line-level HTR are the
Character Error Rate (CER) and the Word Error Rate (WER), borrowed
from the Automatic Speech Recognition field. These metrics indicate
the length-normalised number of elementary editing operations needed
to produce a reference (correctly transcribed) sequence from the HTR
hypothesis, at the character (CER) or word (WER) level. Under the
premise of a line-level formulation, it is generally acknowledged that
these metrics provide a good measure of performance.

Due to recent advances in the field, especially brought about by the
intensive use of deep neural networks, line-level HTR is considered
practically solved, or close to.  Therefore, the field is experiencing
a paradigm shift towards end-to-end full-page scenarios.  In a
page-level application, lines are not given. Instead, images are
usually processed to first extract single lines, under a process
generally known as Layout Analysis (LA).\!%
\footnote{Many present-day HTR systems use simplified forms of LA
  which only focus on detecting the text-lines of each image.  In the
  sequel, the term LA will be used indistinctly to refer to proper LA
  as well as to just line detection.}
Then, each line is transcribed independently with line-level
HTR. Furthermore, some works do not explicitly include any LA step
and aim to obtain the transcription hypothesis by processing whole
pages or paragraphs~\cite{Bluche2016,Coquenet2021}. 

\begin{figure}
\vspace{-0.2em}
\def\s{-0.3ex} \def\ss{+2.6ex} \newcommand{\f}[1]{\textsf{\small #1}}
  ~~~~\fbox{\scalebox{.89}{
  \begin{minipage}[t]{.25\textwidth}
    \f{Two ways of coming}\\[\s]
    \f{at the (archetypes of)}\\[\s]
    \f{Geometrical abstract}\\[\s]
    \f{Quantities: 1. by}\\[\s]
    \f{decomposing Bodies:}
  \end{minipage}
  \begin{minipage}[t]{.26\textwidth}
    1. application of\\[\s]
    metaphysics to\\[\s]
    mathematics.\\[\s]
    2. Method of facilitating\\[\s]
    the Study of mathematics
  \end{minipage}}}~
  \begin{minipage}[t]{0.4\textwidth}
    \emph{Two-columns reference transcript}
  \end{minipage}
  \\[-1.6em]
  \begin{minipage}[t]{\textwidth}
    ~\flushright\emph{Automatic transcript (WER=70.0\%)}~~~~~~
  \end{minipage}\\[-0.6em]
  \raisebox{-17pt}{~~~~\includegraphics[width=0.48\textwidth]{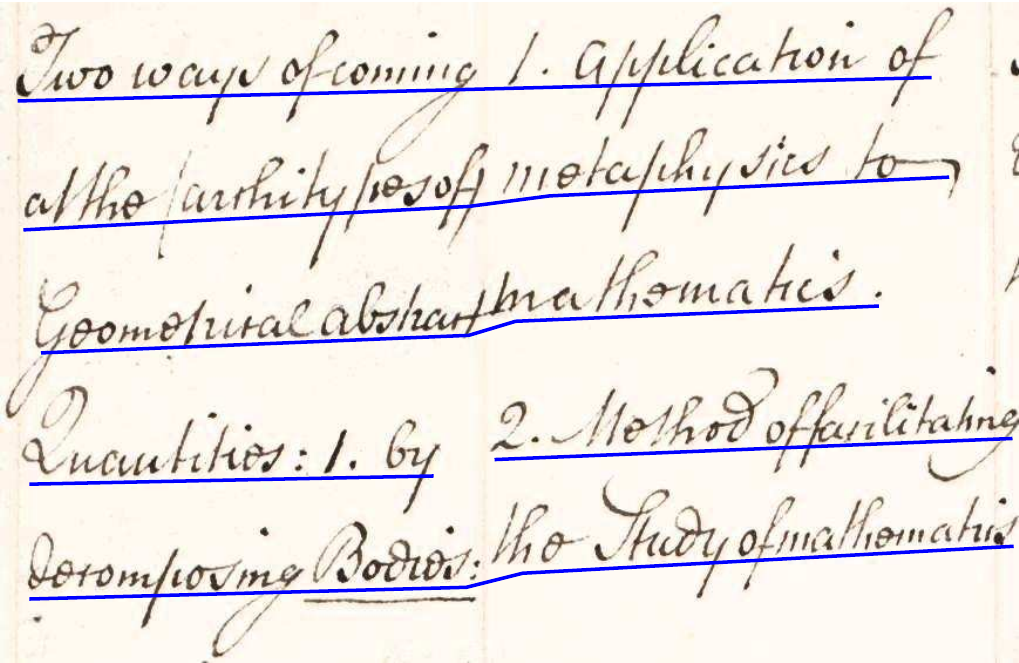}}
  ~\raisebox{-0.3em}{\fbox{\scalebox{.88}{
  \begin{minipage}[b]{.49\textwidth}
    \f{Two ways of coming 1. application of}\\[\ss]
    \f{at the (archetypes of) metaphysics to}\\[\ss]
    \f{Geometrical abstract mathematics.}\\[3ex]
    \f{2. Method of facilitating}\\[-0.2ex]
    \f{Quantities: 1. by}\\[1ex]
    \f{decomposing Bodies: the Study of mathematics}\\[-0.6em]
  \end{minipage}}}}
  \vspace{-0.8em}
  \caption{\label{fig:la-errors-for-htr} Example of misleading WER
    evaluation, caused by wrong reading order due to text-line
    detection flaws.
    While all the words in the automatic transcript are perfectly
    correct, the WER is 70\% (13 matching words, 13 substitution
    errors, 4 insertions and 4 deletions).
    The image is part of a page of the Bentham Papers collection
    (see Sec.\,\ref{sec:ro}).
  }
\vspace{-1em}
\end{figure}

Despite moving from the line-level to the page-level HTR scenario, the
traditional CER and WER metrics are still generally used for
assessment. However, this evaluation protocol is too naive: full-page
real applications do suffer from LA errors which systematically lead
to inconsistencies when evaluating the model using such metrics.
Figure \ref{fig:la-errors-for-htr} shows a real example of this kind
of issues related to LA (see other examples in
Figs.\,\ref{fig:complexExamples},~\ref{fig:splitLines},
and~\ref{fig:evalExample}).  While all the words are perfectly
recognised, the WER is 70\%, which is absolutely misleading.  Clearly,
if this figure is meant to reflect anything, it is a LA problem ---
nothing related with word recognition errors!
This kind of problems become even more insidious in approaches that
bypass the LA step. When researchers were hard-pressed to obtain
acceptable performance values, questioning the traditional evaluation
protocol did not seem relevant.  However, with an increasing number of
effective page-level transcription workflows, we see the need to ask
ourselves about the nature of its evaluation and whether the
traditional line-level evaluation faithfully represents a proper
indicator of page-level transcription performance.

The difficulties underlying the evaluation of page-level HTR results
boil down to a Reading Order (RO)
problem~\cite{clausner2013,Pletschacher2015,%
naoum2019,quiros2022}.
%
A number of recent proposals try to heuristically weight and combine
both word recognition and LA geometric errors into a single scalar
value~\cite{LeifertLGL19,coquenet2023dan}.  Unfortunately, this
hinders the capability to sort out the nature of the corresponding
errors and thereby making a comprehensive, useful assessment.
Here we instead advocate for a two-fold evaluation approach which
decouples the impact of word (and character) recognition errors from
the influence of wrong RO and, furthermore, it is largely agnostic to
geometry-related flaws.

One possibility to assess page-level word recognition accuracy
regardless of RO is to rely on the Bag of Words concept, as proposed
and used in early works by Antonacopoulos, Clausner and
Pletschacher~\cite{Pletschacher2015,Clausner2017,clausner2017a,%
  ClausnerAP19} (see also~\cite{ClausnerPA20}), and later by other
authors~\cite{Strobel2020}.
Here we will argue that a properly defined WER based on the Bag of
Words concept can not trivially consist on just counting how many
words do not appear both in the reference and HTR transcripts.  So we
(re-)define a bag-of-words WER (bWER) so that it becomes faithfully
comparable with the traditional WER and proves to be a very convenient
page-level RO-independent word error metric.

However, the bWER approach does not allow measuring character-level
error, nor it provides the word alignment information needed to
compute RO assessment metrics.  Instead, both word- and
character-level RO-independent recognition accuracy can be precisely
computed using the well-known Hungarian Algorithm
(HA)~\cite{kuhn1955hungarian,burkard2012assignment}.
Here we introduce a regularised version of the HA which provides
HA-based WER values (hWER) that are almost identical to those of bWER
and, moreover, are also practically equal to those of the classical
WER when the reference and HTR transcripts are in the same RO.  In
addition, it further provides the information needed to compute RO
assessment metrics such as the Normalised Spearman's Footrule Distance
(NSFD) \cite{ravi2010,quiros2022}.

In this work, we study all these related approaches to separately
assess at the page-level both the HTR word (and character) recognition
accuracy and the quality of the RO . The problems considered and the
proposed solutions will be presented along with empirical results
obtained on a semi-artificial task, where the typically expected LA
errors and associated RO problems are simulated.  The proposed
assessment methods will be then applied to a series of real page-level
end-to-end HTR experiments, considering both LA-based and holistic
page-level transcription approaches.

Our experiments will show that: i) in the traditional line-level
setting, bWER and WER are typically almost identical; ii) the WER
based on the regularised HA is almost the same as the bWER and both
accurately approach page-level WER in the traditional line-level
evaluation setup; iii) the difference between WER and bWER highly
correlates with the NSFD and is much more efficient than using the HA,
needed to compute the NSFD.

The remainder of this work is structured as follows. Classical WER and
CER measures are reviewed in Section \ref{sec:wcer}; the RO problem
and the NSFD measure are discussed in Section \ref{sec:ro}; the
proposed bWER and hWER metrics are described in Section \ref{sec:bow}
and \ref{sec:hungarian}, respectively; and a summary of the
different metrics considered is provided in Section
\ref{sec:summMetrics}. Then, simulated and real experiments are
reported and analysed in Section \ref{sec:simuExp} and
\ref{sec:realEval}, respectively. We close the article by outlining
related works in Section \ref{sec:related} and concluding in Section
\ref{sec:concl}.  Finally, \ref{ap:ex} presents detailed examples of
the computation of the different metrics proposed and
\ref{ap:software} provides details for public access to the datasets
and software tools used and developed in this work.

\vspace{-0.4em}
\section{Word \& Character Error Rates Based on the Edit Distance }
\label{sec:wcer}
\vspace{-0.2em}


Traditional HTR assessment is based on line-level WER and CER.  As
commented above, this ignores possible line detection and/or
extraction errors made by the LA stage in real automatic transcription
tasks.
This section reviews this evaluation approach, as an introduction to
the forthcoming sections, where we propose new approaches for fair
page-level end-to-end HTR assessment.

\vspace{-0.4em}
\subsection{Edit Distance, WER and CER for Word Sequences}
\label{sec:basicED}
\vspace{-0.1em}

Let the word sequences $x=x_1,\dots,x_{|x|}$ and $y=y_1,\dots,y_{|y|}$
be a reference text and a HTR hypothesis, respectively.
The word \emph{edit distance} from $x$ to $y$, $\edit(x,y)$, is the
minimum number of word insertion, substitution and deletion edit
operations that transform $x$ into $y$~\cite{wagner74}.
Edit operations define a ``trace'' or alignment between word instance
positions of $x$ and $y$, which may be formulated in several
equivalent ways.  Here we loosely follow the work of Marzal and
Vidal~\cite{marzal93} and define an alignment $\mathcal{A}(x,y)$ as a
sequence of ordered pairs of integers (word indices), $(j,k)$,
$1\!\leq\!j\!\leq\!|x|,~1\!\leq\!k\!\leq\!|y|$, such that for every
two distinct pairs
$(j,k),(j',k')\in\mathcal{A}(x,y),~j\!<\!j'\Leftrightarrow k\!<\!k'$.
In what follows, word alignments which fulfil this
\emph{sequentiality constraint} will be denoted as
$\mathcal{T}(\cdot,\cdot)$, leaving the notation
$\mathcal{A}(\cdot,\cdot)$ only for unconstrained alignments.

$\mathcal{T}(x,y)$ can be conveniently extended to explicitly
represent word insertions and deletions.  To this end, a \emph{dummy}
position, denoted by $\epsilon$, is assumed in both $x$ and $y$ which
points to the \emph{``empty word''}, $\lambda$; that is,
$x_\epsilon=y_\epsilon\eqdef\lambda$.
The edit distance from $x$ to $y$ is thus formally defined as:
\vspace{-0.4em}
\begin{equation}\label{eq:editDist}
  \edit(x,y) ~=~ \min_{\mathcal{T}(x,y)}
   \sum_{(j,k)\in\mathcal{T}(x,y)}\!\!\! \updelta(x_j,y_k)~~~~~~~~
\vspace{-0.4em}
\end{equation}
where $\updelta(a,b)$ is defined to be $1$ if $a\neq b$ and $0$ otherwise.%
With these editing costs, it is often called Levenshtein distance.
For the above sequentiality constraint to still be meaningful, we
assume that the predicate $j\!<\!j'$ is \emph{true} for any $j,j'$
such that $j$ or $j'$ are $\epsilon$.

By analysing the pairs in the optimal trace $\mathcal{T}(x,y)$, the
sum in Eq.\,\eqref{eq:editDist} can be decomposed into separate counts for
insertions, substitutions and deletions; i.e., ~$\edit(x,y)=i+s+d\,$.
%
Example~1 in \ref{ex:1} illustrates the computation of the word edit
distance and the corresponding trace%
\footnote{\label{foot:tok}To avoid nonessential complications such as
  (language-dependent) tokenization and capitalisation, any character
  sequence delimited with withe space is considered a ``word''.
  Therefore: $\updelta$(\textsf{\scriptsize
    be},\,\textsf{\scriptsize be,}) $=$ $\updelta$(\textsf{\scriptsize
    be},\,\textsf{\scriptsize be:}) $=$ $\updelta$(\textsf{\scriptsize
    The},\,\textsf{\scriptsize the}) $=1$.
} %
for $x=\,\,$\textsf{\small ``To be or not to be, that is the question''}
and $y=\,\,$\textsf{\small ``to be oh! or not to be: the question''},
with $i=1, s=2, d=2\,$ and $\,\edit(x,y)=5$.



The WER of $y$ with respect to $x$ is defined as the edit distance,
normalised by the length of the reference text,%
\footnote{Note that, defined in this way, it may happen that
  $\WER(x,y)\!>\!1$, which prevents WER to be properly interpreted as
  an error probability.  For the same reasons a Word Accuracy can not
  be defined just as $1-\WER$. The Normalised Edit
  Distance~\cite{marzal93,vidal95} would overcome these drawbacks but,
  following time-honoured tradition in ASR and HTR alike, we stick with
  the conventional normalisation by the length of the reference
  sequence.
} %
$n=|x|$:
\vspace{-1.2em}
\begin{equation}\label{eq:lineWER}
  \WER(x,y) ~=~      \frac{\edit(x,y)}{n}
            ~\equiv~ \frac{i+s+d}{c+s+d}~~~~~~~
\vspace{-0.2em}
\end{equation}
where $c$ is the number of correct words (those which do not need
editing).
In Example~1~(\ref{ex:1}), $\WER(x,y)$ $=(1+2+2)/(6+2+2)=5/10=50\%$.

The CER is defined similarly, by just assuming that $n$ is the total
number of characters in $x$ and $i,s,d,c$ are \emph{character}, rather
than \emph{word} edit operations and correct matching counts.  

\vspace{-0.2em}
\subsection{Traditional, Line-based Page Level WER and CER}
\label{sec:traditionalWerCer}
\vspace{-0.1em}

                                          \enlargethispage{1.5em} 

Let $\mathcal{I}$ be a text image and $X$ the reference GT transcript
of $\mathcal{I}$. Let $Y$ be the transcription hypothesis provided by
an HTR system for $\mathcal{I}$.  Both $X$ and $Y$ are made up of the
same number $M$ of individual text-lines $x^1,x^2,\dots x^M$ and
$y^1,y^2,\dots y^M$, respectively, where each text-line is a sequence
of words.
Each pair of text-lines $x^\ell$ and $y^\ell$ are transcripts of the
same image line, which is simply denoted as
$\ell,1\!\leq\!\ell\leq\!M$.
%
%
In the traditional setting, page-level WER is then computed as:
\vspace{-0.3em}
\begin{equation}\label{eq:pageLineWER}
  \WER(X,Y) ~=~ \frac{\sum_{\ell=1}^M \edit(x^\ell,y^\ell)}{N}
       ~\equiv~ \frac{\sum_{\ell=1}^M(i_\ell+s_\ell+d_\ell)}
                     {\sum_{\ell=1}^M(c_\ell+s_\ell+d_\ell)}
\vspace{-0.3em}
\end{equation}
where $N=|X|$ is the total number of word instances of $X$.


Another way to compute $\WER(X,Y)$ is to concatenate all the $M$ lines
of $X$ and $Y$ in any arbitrary order (the same order for $X$ and $Y$)
and directly compute the edit distance between the concatenated
texts.  Except for small possible differences in the text-line
boundaries, the editing operations obtained by this computation will
be essentially the same as those involved in the $M$ edit distances
$\edit(x^\ell,y^\ell), 1\!\leq\!\ell\!\leq\!M$ of
Eq.\,\eqref{eq:pageLineWER}.  Therefore:
%
\begin{equation}\label{eq:fullPageWER}
  \WER(X,Y) ~\approx~ \frac{\edit(X,Y)}{N}
            ~\equiv~  \frac{i'+s'+d'}{c'+s'+d'}
\end{equation}
where $i',s',d'$ and $c'$ are now counts of word edit operations and
matchings involved in the computation of $\edit(X,Y)$ for the
\emph{whole} texts $X$ and $Y$.

As in Sec.\,\ref{sec:basicED}, the CER is defined similarly by just
assuming that $N$ is the total number of characters in $X$ and
$i,s,d,c,i',s',d',c'$ are \emph{character}, rather than \emph{word}
edit operation and correct matching counts.


\vspace{-0.4em}
\subsection{Page-level End-To-End Assessment Using Traditional WER and CER }
\label{sec:traditEndToEnd}
\vspace{-0.1em}

In a realistic scenario, image-lines may be given for the GT reference
transcript, $X$.  But these lines may not correspond one-to-one with
lines automatically detected in the text image $\mathcal{I}$.
Moreover, the number of text-lines in $X$ and $Y$ might be different.

To overcome this hurdle, it is often ignored that the lines of $Y$ may
\emph{not} be in the same \emph{reading order} (RO) as those of $X$
and the WER is thus naively computed for the whole texts in $X$ and
$Y$ as in Eq.\,\eqref{eq:fullPageWER}.
This is the approach often followed in experiments which aim to
provide end-to-end performance assessment such
as~\cite{Bluche2016} and~\cite{sanchez2019} (Sec.8,
Test-B2).

\vspace{-0.3em}
\section{The Reading Order Problem}\label{sec:ro}
\vspace{-0.3em}

The RO of a sequence of words $W=w_1,\dots,w_n$ is just the linear
sequence $1,\dots,n$ of the \emph{positions} of these words in $Z$.
Loosely speaking, two transcripts $X$ and $Y$ of an image
$\mathcal{I}$ are said to be in a similar RO if a sequential,
monotonous correspondence (i.e., a trace) exists between the positions
of the matching words of $X$ and $Y$.
Note that this applies to documents written in occidental or
latin-derived left-to-right writing style, as well as to other
scripts where writing follows right-to-left or top-to-bottom
directions.

\begin{figure}[htbp]
  \centering
  \includegraphics[height=.36\linewidth]{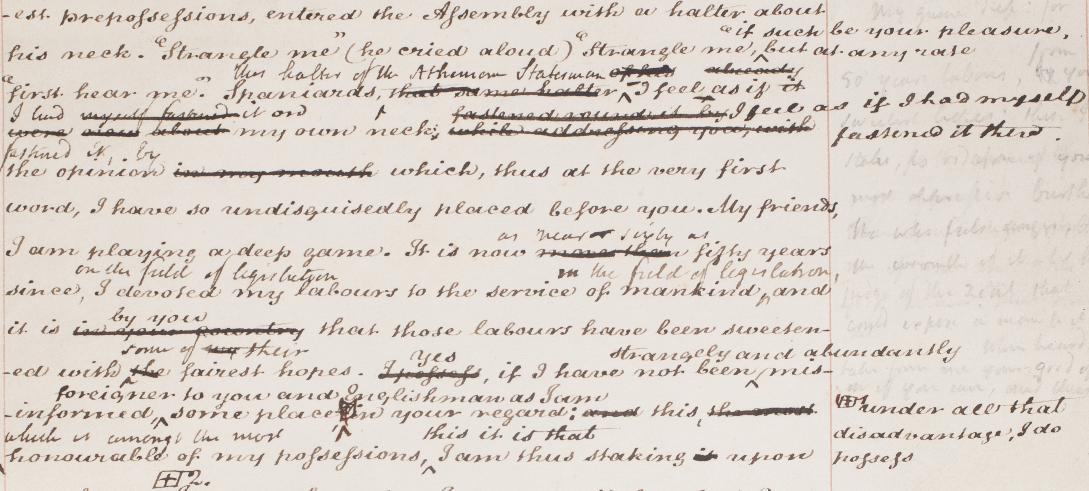}\\[0.35em]
  \includegraphics[height=.26\linewidth]{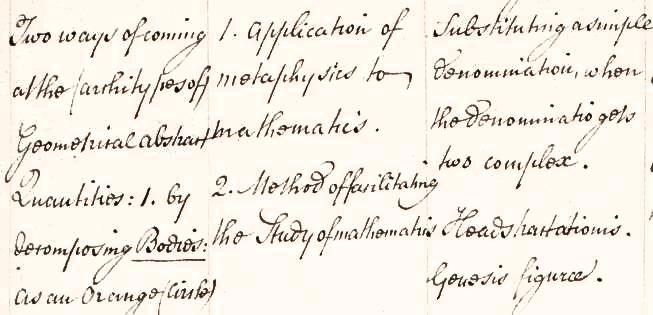}%
  \hspace{.5em}%
  \includegraphics[height=.26\linewidth]{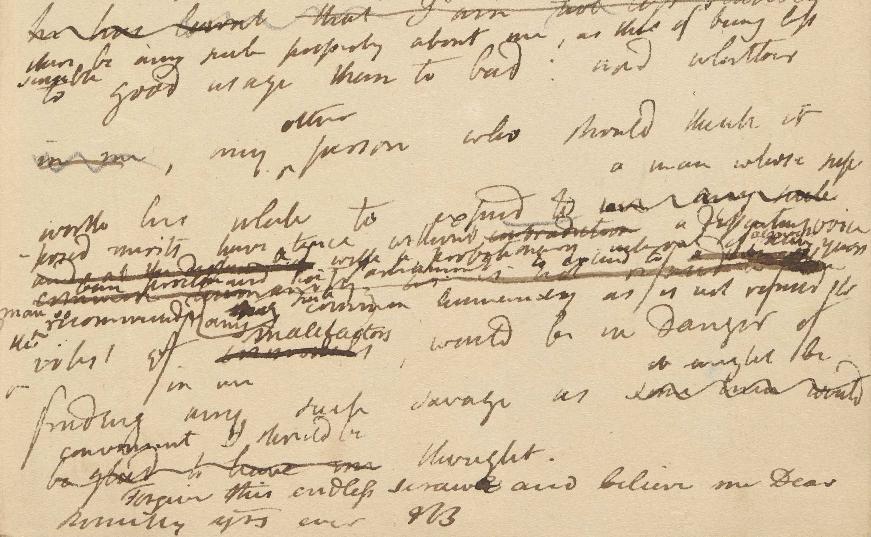}
  \vspace{-0.5em}
  \caption{Examples of frequent reading order issues, from the Bentham
    Papers collection.}
  \label{fig:complexExamples}
\vspace{-0.5em}
\end{figure}

As noted in Sec.\,\ref{sec:intro}, assuming that reference and
hypothesis transcripts are in similar RO is generally unrealistic.
This is particularly the case in many historical handwritten text
images such as those shown in Fig.\,\ref{fig:complexExamples}.%
\footnote{From the Bentham Papers collection.  See, e.g.:
  \url{http://prhlt-kws.prhlt.upv.es/bentham}}
%
%
%
If Eq.\,\eqref{eq:fullPageWER} of Sec.\,\ref{sec:traditEndToEnd} is
applied in this scenario, the resulting WER figures will reflect an
uncontrolled combination of actual word recognition failures and
errors due to inaccurate RO generally due to poor LA.
%

On the other hand, it is important to realise that
the RO provided by reference transcripts and/or other layout GT
annotations is generally only one among several possible RO
annotations which would be all correct.
Therefore, mixing RO and word recognition errors into a single
assessment measure (as in~\cite{coquenet2023dan,LeifertLGL19}) does
not seem the best idea for understanding which are the inner issues of
an end-to-end full-page HTR system.

These facts lead us to propose a two-folded evaluation approach which
completely decouples the RO from word recognition errors, while also
providing a simple, comprehensive picture of the end-to-end system
performance.

Research on RO has some tradition for printed
documents~\cite{clausner2013,naoum2019,lee2021}.
%
More recently, RO analysis has also been considered for handwritten
documents, where RO issues are specially relevant.  In the work of
Quiros and Vidal~\cite{quiros2022}, effective methods to
learn line RO in handwritten text images from examples are proposed
and empirically assessed.
%

\subsection{Assessing Reading Order: Normalised Spearman's Footrule Distance}
\label{sec:roAssess}

RO assessment issues are discussed in~\cite{quiros2022}, where two
metrics are finally proposed and used in the experiments: the
Kendall's Tau rank distance (also called bubble-sort
distance)~\cite{kendall1938} and the Normalised Spearman's Footrule
Distance (NSFD)~\cite{ravi2010}.  Here we adopt the latter because it
measures not only how many elements are not placed in the correct
position within the expected order, but also how far these elements
are from their correct positions.  Thereby it provides reasonable
estimates of the human effort that would be needed to render a
sequence of elements in a correct order given by a reference sequence.
The NSFD can be defined as:
\begin{equation}\label{eq:nsfd}
  \rho(X,Y) ~=~ \frac{1}{\lfloor\frac{1}{2}N^2\rfloor}
                \!\sum_{(j,k)\in\mathcal{A}(X,Y)}\!\!\! |j-k|
\vspace{-0.5em}
\end{equation}
where $\mathcal{A}(X,Y)$ is an alignment between the reference text
$X$ and the HTR hypothesis $Y$, and $N=\max(|X|,|Y|)$ is the number of
words of the longest text.
Note that the alignment $\mathcal{A}(X,Y)$ does \emph{not} need to
fulfil the sequentiality constraint used in Sec.\,\ref{sec:basicED}
to define the word edit distance.  In what follows, we
assume that the alignment used in Eq.\,\eqref{eq:nsfd} will be
provided by the methods discussed in Sec.\,\ref{sec:hungarian}.
Example~2 in \ref{ex:2} illustrates the computation of the NSFD for
$X\!=\,\,$\textsf{\small ``To be or not to be, that is the question''}
and $Y\!=\,\,$\textsf{\small ``The big question: to be or not to be''},
with $\rho(X,Y)=27/50~ (54\%)$.

From a user point of view, insertions and deletions do not typically
affect the RO in a substantial way. Therefore, in Eq\,\eqref{eq:nsfd}
we just assume that $|j-\epsilon|=|\epsilon-k|\,\eqdef\,1~\forall
j,k$.
However, insertions and deletions may indirectly affect significantly
the result of Eq\,\eqref{eq:nsfd}, because of the contribution of
subsequent values of $|j-k|$.  This is illustrated in Example~2 as
well, along with the approach we propose to circumvent this problem by
just renumbering the positions of words of $Y$ and/or $X$ according to
the inserted or deleted words specified in $\mathcal{A}(X,Y)$.

%

\vspace{-0.6em}
\section{Bag of Words WER}\label{sec:bow}
\vspace{-0.2em}

In Sec.\,\ref{sec:wcer}, $X$ and $Y$ were considered sequences where
the order of text-lines and words is relevant for computing word
errors. However, in page-level performance assessment, once we have a
specific metric to measure RO, it is desirable to largely ignore the
order of words in $X$ and $Y$ to measure word recognition performance.

A simple way to achieve this goal is to rely on the ``Bag of Words''
concept, as discussed in Sec.\,\ref{sec:intro}.
%
To this end, $X$ and $Y$ are now considered multi-sets (or ``bags'')
of words and the number of instances of each word can be used to
compute a metric which is fairly closely related to the WER.

Let $V_X$ and $V_Y$ be the respective sets of different words
(vocabularies) of $X$ and $Y$, and $V=V_X\cup V_Y$.  For each word
$v\in V$ let $f_X(v)$ and $f_Y(v)$ be the number of instances of $v$
in $X$ and $Y$, respectively. The ``bag of words distance'' between
$X$ and $Y$ is defined as:
\vspace{-0.2em}
\begin{equation}\label{eq:BoWdist}
  \mathrm{B}(X,Y) = \sum_{v\in V}|f_X(v)-f_Y(v)|~~~~~
\vspace{-0.5em}
\end{equation}
\noindent
Then, if $N$ is the number of words in the reference $X$, a simple
``BoW WER'' can be rather naively defined as:
\vspace{-0.2em}
\begin{equation}\label{eq:betaWER}
  \beta\!\WER(X,Y)~=~\frac{B(X,Y)}{N} ~\equiv~
                     \frac{1}{N}\sum_{v\in V}|f_X(v)-f_Y(v)|~~~
\vspace{-0.3em}
\end{equation}

As defined in Eq.\,\eqref{eq:BoWdist}, $\mathrm{B}(X,Y)$ is the number
of word instances of $X$ which fail to appear in $Y$ plus the number
of word instances in $Y$ which are not in $X$.  This can be properly
interpreted in terms of editing operations just as the total number of
word insertions and deletions that would be needed to transform $X$
into $Y$, \emph{without allowing for word substitutions}.

In the classical WER, a combined deletion and insertion pair of edit
operations can be achieved by a single substitution.  So, if $X$ and
$Y$ are in the same RO, the bag of words distance will always be
larger than or equal to the corresponding word edit distance; that is,
$\mathrm{B}(X,Y)\geq\edit(X,Y)$.
If word substitution were allowed, many pairs of the $\mathrm{B}(X,Y)$
insertions and deletions could be advantageously exchanged by single
substitutions.  In the best case, the number of these word
substitutions would be exactly $\mathrm{B}(X,Y)/2$.
However, if $|X|\neq|Y|$, it is unavoidable that a number of words
$b=\big||X|-|Y|\big|$ have to be actually deleted or inserted, without
any possible pairing for interpretation as single substitutions.  We
will say that these insertions or deletions are \emph{unavoidable}.

Therefore, to define a ``bag of words WER'' which can be fairly
compared with the traditional WER, we assume that each
insertion/deletion pair, except those unavoidable, is equivalent to a
single substitution.
Formally speaking, the above definition of bag of words distance needs
to be revamped into
$\mathrm{B}'(X,Y)=b+\lfloor(\mathrm{B}(X,Y)-b)/2\rfloor$.  Since
$\mathrm{B}(X,Y)-b$ is always even, the \emph{bag of words WER} is
thus defined as:
\vspace{-0.3em}
\begin{equation}\label{eq:werBoW}
  \bWER(X,Y) ~=~ \frac{\mathrm{B}'(X,Y)}{N}
             ~=~ \frac{1}{2N}\,
         \biggl(\,\big|\,N-|Y|\,\big|~+~\sum_{v\in V}|f_X(v)-f_Y(v)|\,\biggr)~~~
\vspace{-0.5em}
\end{equation}
%
%
%

Through the computation of Eq.\,\eqref{eq:werBoW}, the \emph{number}
of word insertions, deletions and (implicit) substitutions can be
easily derived, even though which specific words are involved in the
different operations remain unknown.
This becomes a significant drawback, because it prevents to derive any
kind of word-to-word or position-to-position alignment that could be
used to compute the NSFD or any other metric to assess RO mismatch.
For the same reason, a CER associated with bWER can neither be
properly computed.  (Note that a ``bag of characters'' error rate
would be overtly deceptive, and therefore is not an option).
%
The examples in \ref{ex:3} illustrate the computation of $\beta\,$WER
and the reformulated version here proposed bWER (Eq.\,\eqref{eq:werBoW}),
along with their relation with the classical WER.

%
%
%

It is important to note that the WER is based on sequentially
constrained alignments (see Section \ref{sec:basicED}), while the bWER
does not. Therefore, bWER can be (much) lower than WER, especially if
the RO of $X$ and $Y$ are very different.
Even without the RO issue, the bWER can underestimate word errors.
Example 3a in \ref{ex:3} shows a simple case of this. However, based
on empirical evidence presented in Sections \ref{sec:simuExp} to
\ref{sec:realEval}, these cases are rare in practice.
The page-level bWER (Equation \eqref{eq:werBoW}), therefore, becomes a
good approximation to the corresponding WER in traditional
experimental settings where RO is not an issue.  This is interesting
because bWER is much simpler and cheaper to compute than WER.

\section{CER, WER and NSFD Based on Bipartite Graphs and the 
                                              Hungarian Algorithm}
\label{sec:hungarian}

As discussed above, determining RO-independent word and character
recognition accuracy at the full-page level, requires words and/or
word positions from the reference transcript $X$ to be freely aligned
or paired with corresponding words of the transcription hypothesis
$Y$.
Edit distance computation provides word alignments (traces) as a
byproduct, but the trace sequentiality restriction leads to alignments
which lack the freedom needed for RO-independent word pairing.
A proper formulation of the required kind of word alignments is given
by the so-called \emph{``minimum-weight matching or assignment
  problem''}~\cite{burkard2012assignment}.

Let $G=(V,E)$ be a \emph{bipartite graph}, where the set of
nodes $V$ is composed of two disjoint subsets $A,B$, $A\cup B\!=\!V$,
$A\cap B\!=\!\emptyset$, and the set of edges $E$ is a subset of
$V\!\times V$ such that
$(u,v)\in\!E\Rightarrow u\in\!A\,\wedge\,v\in\!B$.
A \emph{matching} $\mathcal{M}\!\subset\!E$ is a set of pairwise
non-adjacent edges; that is, no two edges share a common node.
A node is matched if it is an endpoint of one of the edges in the
matching. Otherwise, the node is unmatched.
$\mathcal{M}$ is said to be \emph{maximum} if it contains the largest
possible number of edges
%
and it is a \emph{perfect matching} if all the vertices of the
graph are matched.
Every perfect matching is also maximum.
%
A bipartite graph $G=(V,E)$ is \emph{weighted} if a real-valued
weight $g(u,v)$ is assigned to each edge $(u,v)\in\!E$.  Then, the
weight of a matching $\mathcal{M}$ is the sum of the weights of the edges
in $\mathcal{M}$.
Given a weighted bipartite graph, the assignment problem is to find a
perfect matching with minimum weight.
An efficient solution to this problem is provided by the
\emph{Hungarian Algorithm} (HA)~\cite{kuhn1955hungarian}.
%

\medskip
In our HTR assessment task, $A$ and $B$ are, respectively, the word
instances of the reference transcript $X$ and the HTR hypothesis $Y$
of a page image; $E\!=\{\!(X_j,Y_k), 1\!\leq\!j\!\leq\!|X|,
1\!\leq\!k\!\leq\!|Y|\}$ is the set of all pairs of word instances in
$X$ and $Y$, and
the weight $g(X_j,Y_k)$ is the character edit distance between the
$j$-th word of $X$ and the $k$-th word of $Y$.
Word insertions and deletions are represented by assignments to ``dummy''
nodes, which represent the empty word $\lambda$.  These nodes need to
be added to \emph{both} sets, not only because in general
$|X|\neq|Y|$, but also because we need to simultaneously support
\emph{both} \emph{insertions} and \emph{deletions} for any given pair
of transcripts.
The cost of an edge connecting a dummy node with a word $v$ is thus
defined as $g(v,\lambda) = g(\lambda,v)=\frac{1}{2}|v|$, where $|v|$
is the number of characters of $v$ and, as in Sec.\,\ref{sec:bow}, the
factor $\frac{1}{2}$ is introduced to balance the cost of a word
substitution with that of an equivalent combined word insertion and
deletion.

\begin{figure}[htb]
\centering
\!\!\!\!\!\includegraphics[height=.35\linewidth]{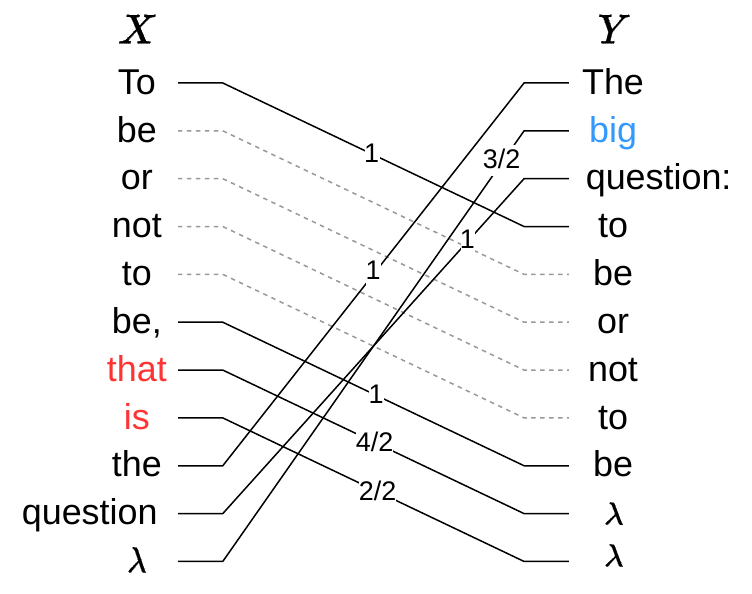}%
\vspace{-0.8em}
\caption{\label{fig:hung} Assignment obtained by the Hungarian
  algorithm for a bipartite graph corresponding to the word sequences
  $X$ and $Y$.  Dotted edges have a null cost and coloured words are
  insertions or deletions.  For this assignment,
  $ \mathrm{d_h}(X,Y)=1+1+4/2+2/2+1+1+3/2=8.5$.}
\vspace{-0.5em}
\end{figure}

The assignment problem is to pair each (maybe empty) word instance of
$X$ with a (maybe empty) word instance of $Y$ so that the sum of
character edit distances between the paired words is minimum.
Therefore, the HA yields what could be called ``HA Character Edit
Distance'':
%
\begin{equation}\label{eq:hungCER}
  \mathrm{d_h}(X,Y) ~=~ \min_{\mathcal{A}}
  \!\!\sum_{(j,k)\in\mathcal{A}(X,Y)}\!\! g(X_j,Y_k)\,;
\end{equation}
%
%
Fig.\,\ref{fig:hung} illustrates all the above concepts for a pair of
word sentences.
\vspace{-0.3em}

The optimal alignment $\hat{\mathcal{A}}(X,Y)$ associated with
Eq.\,\ref{eq:hungCER} is a set of pairs $(j,k)$,
$1\!\leq\!j\!\leq\!|X|\!=\!N$, $1\!\leq\!k\!\leq\!|Y|$, along with two
additional sets of pairs of the form $(j,\epsilon)$ and $(\epsilon,k)$
to account for word deletions and insertions, respectively.
Let $D$ be the number of these dummy pairs in $\hat{\mathcal{A}}(X,Y)$
and, as in Eq.\,\eqref{eq:werBoW}, let $b=\big|\,|X|-|Y|\,\big|$.
Since both insertions and deletions are allowed in
$\hat{\mathcal{A}}(X,Y)$, $D\geq b$.  So, as in the case of
Eq.\,\eqref{eq:werBoW} for the bWER, the (now typically few) $D-b$ excess
pairs of insertions and deletions, can be interpreted as single
substitutions.
Then, the ``HA WER'' (hWER) can be defined as:
\vspace{-0.5em}
\begin{equation}\label{eq:hungWER}
  \hWER(X,Y) ~=~ \frac{1}{N}\!
  \!\!\!\sum_{\:\:\:(j,k)\in\hat{\mathcal{A}}(X,Y)}
  \!\!\!\updelta(X_j,Y_k) ~~-~~ \frac{(D - b)}{2N}
\vspace{-0.2em}
\end{equation}
where $\updelta(\cdot,\cdot)$ is the $0/1$ function introduced in
Sec\,\ref{sec:basicED}.
Also using $\hat{\mathcal{A}}(X,Y)$, the NSFD $\rho(X,Y)$ can be
computed straightaway as in Eq.\,\eqref{eq:nsfd}.


To compare hWER with bWER, note that the optimisation of
Eq.\,\eqref{eq:hungCER} ensures a word alignment with minimum sum of
\emph{character edit distances} between the paired words.  But this
alignment may not always lead to a minimum \emph{word edit distance}.
Thus, while it can be easily shown that
$\bWER(X,Y)\leq\hWER(X,Y)~\forall X,\!Y\!$,\, the strict equality may
not hold in some cases.

%

The examples in \ref{ex:4} further illustrate the
computation of \,hWER for the more realistic texts used in
Example~3. 
It is worth noting that the values of hWER in these examples are
identical to the corresponding bWER values of Example~3~(\ref{ex:3}).

\medskip
When multiple instances of some word exist in $X$ and/or in $Y$, as in
the examples of \ref{ex:4}, the HA is free to pair any matching
instances, as long as the values of $\mathrm{d_h(X,Y)}$ are the same.
In other words, there may be multiple alignments which provide the
same optimal result for Eq\,\eqref{eq:hungCER} and the HA has no means
to decide which one would be more consistent with the positions of
these words in the RO of the compared texts.

This is discussed in detail in \ref{ex:5} for one of the examples of
\ref{ex:4}.  Because of unlucky tie breaks, the NSFD between two
example sentences $X$ and $Z$ which are almost in the same RO is
$\rho(X,Z)\!=\!13.3\%$.  However, if ties are broken more favourably
(and in a more natural way), the resulting NSFD is
$\rho(X,Z)=1/\lfloor14^2/2\rfloor = 1.0\%$, which much better reflects
the very minor RO discrepancy between $X$ and $Z$.

%
\medskip
To avoid this kind of ties, we propose to regularise the HA cost with
a term which measures the contribution of each pairing to increase the
NSFD.  That is, we propose changing Eq.\,\eqref{eq:hungCER} into:\\[-0.5em]

\vspace{-1.8em}
\begin{equation}\label{eq:hungRegulCER}
  \mathrm{d_h}(X,Y) ~=~ \min_{\mathcal{A}}
  \!\!\sum_{(j,k)\in\mathcal{A}(X,Y)}\!\! \biggl(g(X_j,Y_k) 
  ~+~ \gamma\,\frac{|j-k|}{N} \,\biggr) 
\end{equation}
%
where $\gamma$ is the regularisation factor and, as in Eq.\eqref{eq:nsfd},
it is assumed that $|j-\epsilon|=|\epsilon-k|=1~\forall j,k$.

If $\gamma$ is close to $0$, the HA will just behave as usual,
yielding hWER values very close or identical to those of bWER, but
alignments $\hat{\mathcal{A}}(\cdot,\cdot)$ not ideal for assessing RO
discrepancies.
On the other extreme, for large $\gamma$ the HA will tend to provide
alignments which do not change word order; that is, alignments close
to the sequential trace $\mathcal{T}(\cdot,\cdot)$ of the traditional
edit distance (cf. Eq.\eqref{eq:editDist}), with NSFD values close to
$0$.
For small values of $\gamma$ it is expected that the hWER result
provided by Eq,\,\eqref{eq:hungWER} and Eq.\,\eqref{eq:hungRegulCER}
will be very close or identical to those obtained with $\gamma=0$; but
the alignment $\hat{\mathcal{A}}(\cdot,\cdot)$, when used in
Eq.\,\eqref{eq:nsfd}, will result is NSFD values which more fairly
reflect RO discrepancies.


\medskip
To define a proper ``HA character error rate'' (hCER), note that the
HA score $\mathrm{d_h}(X,Y)$ is not directly suitable because of the
regularisation and the special treatment of word insertions and
deletions.
However, a simple approximation can be easily computed as
$\hCER(X,Y)\,\eqdef\,\CER(X,\tilde{Y})$, where $\CER(\cdot,\cdot)$ is
the standard character error rate (see
Sec.\,\ref{sec:traditionalWerCer}) and $\tilde{Y}$ is obtained by
reordering the word hypothesis $Y$ according to the optimal alignment
of Eq.\,\eqref{eq:hungRegulCER}.
%
The values obtained in this way for the examples in \ref{ex:4} are:
$~\hCER(X,Y)=8.1$, $~\hCER(X,Z)=16.1$,

\section{Summary of The Different Metrics Proposed}\label{sec:summMetrics}
\vspace{-0.2em}

This section summarises the properties of the most important
evaluation metrics discussed above.  In all the cases, it is assumed
that $X$ is a full-page reference transcript, with $N$ running words,
and $Y$ a corresponding HTR hypotheses with $O(N)$ running words.
%

\begin{description}\itemsep=-0em
\item $\WER(X,Y)$: The traditional Word Error Rate, defined in
  Eqs.\,(\ref{eq:editDist}), and\,(\ref{eq:fullPageWER}), with a
  computational cost in $O(N^2)$.  If $Y$ is in the same RO as $X$,
  the WER just measures the word recognition error rate.  Otherwise,
  this metric is expected to grow monotonically with the amount of RO
  mismatch between $X$ and $Y$, with an offset that reflects the
  actual word recognition failures.  This offset can accurately be
  estimated by the bWER or the hWER.
\item $\beta\!\WER(X,Y)$: An early, naive notion of ``bag of words
  error rate'' defined as $\mathrm{B}(X,Y)/N$, where $\mathrm{B}(X,Y)$
  measures text discrepancies in terms of only word insertions and
  deletions (Eq.\,\eqref{eq:BoWdist}).  It can be computed in $O(N)$
  time.  When $Y$ is in the same RO as $X$, the classical WER yields
  (much) lower values than the $\beta$WER, but if the RO is very
  different, WER is expected to be much larger.  The use of this
  metric is, therefore, not appealing.
\item $\bWER(X,Y)$: A redrafted version of $\beta$WER, given in
  Eq.\,\eqref{eq:werBoW}, which exactly estimates how many word
  insertions and deletions can be equivalently resolved with word
  substitutions.  It can be computed in $O(N)$ time.  When $Y$ is in
  the same RO as $X$, it is expected to yield values which are only
  slightly lower than those of the classical WER but, in contrast to
  WER, it is completely insensible to RO mismatch.  A drawback of this
  metric is that it does not provide any word-to-word alignment,
  thereby preventing to compute a character error rate or to be used
  as a basis to estimate a RO mismatch metric.
\item $\hWER(X,Y)$: The ``Hungarian Algorithm Word Error Rate'',
  defined in Eq.\,\eqref{eq:hungWER} based on a RO-independent word
  alignment obtained as a byproduct of computing
  Eq.\,\eqref{eq:hungRegulCER}.  Its computational cost is $O(N^3)$.  In
  terms of word error rate, hWER is almost identical to bWER, but it
  may provide slightly higher values than bWER in some cases.
  In contrast with bWER, hWER does provide word alignments which allow
  computing a character error rate and can be used to estimate a RO
  mismatch (with the NSFD, e.g.).
\item $\rho(X,Y)$: Normalised Spearman Footrule Distance (NSFD),
  defined in Eq.\,\eqref{eq:nsfd} to explicitly estimate the amount of
  RO mismatch between $X$ and $Y$.  It requires a word-to-word
  alignment which is assumed to be available as a byproduct of
  computing the hWER.  Its computational cost is $O(N)$, but taking
  into account the cost of obtaining the required alignment, the
  overall cost is $O(N^3)$.  The values of NSFD are expected to grow
  monotonically with the degree of RO mismatch.  It is also expected
  that these values be closely correlated with the values of the
  classical WER, after discounting the offset due to actual word
  recognition errors which, as previously mentioned, can be accurately
  estimated by the bWER or the hWER.
  \vspace{-0.3em}
\end{description}

\section{Simulation Experiments}\label{sec:simuExp}
\vspace{-0.2em}

A first series of experiments were carried out to check and
empirically analyse the properties of the proposed metrics under
controlled conditions.
To this end a simple HTR dataset was adopted and real full-page HTR
transcription results were artificially altered in order to simulate
typical conditions that are expected to affect the different evaluation
results.
%
%

\vspace{-0.2em}
\subsection{A Basic Dataset for Testing Different Assessment Approaches}
\label{sec:dataset}

The well known and widely used ICFHR14 dataset was adopted.
This is a small subset of selected manuscripts from the Bentham Papers
collection,\!%
\footnote{The full collection (searchable using PrIx~\cite{toselli2019a})
  is here: \url{http://prhlt-kws.prhlt.upv.es/bentham}} %
mostly written by the English philosopher and reformer Jeremy Bentham.\!%
\footnote{\url{http://blogs.ucl.ac.uk/transcribe-bentham/jeremy-bentham}} %

\begin{figure}[htb]
  \centering
  \includegraphics[height=.35\linewidth]{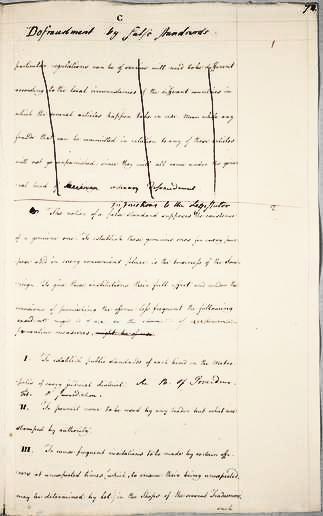}%
  \hspace{.5em}%
  \includegraphics[height=.35\linewidth]{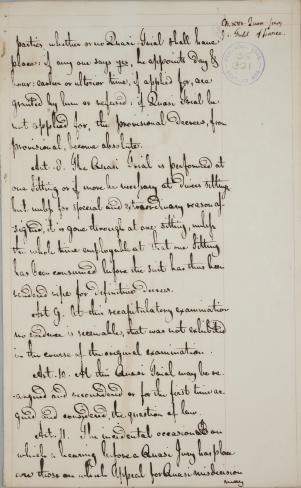}%
  \hspace{.5em}%
  \includegraphics[height=.35\linewidth]{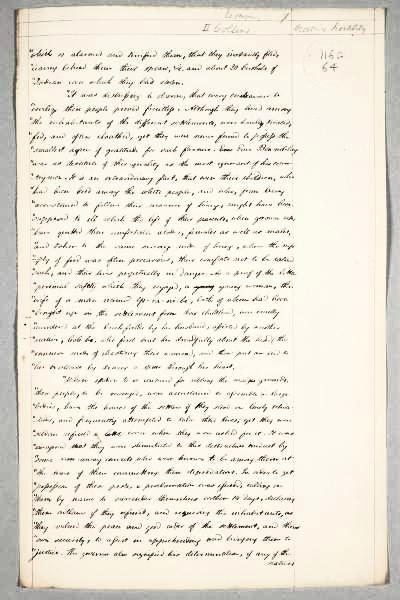}
  \vspace*{-0.7em}
  \caption{Page images of the ICFHR14 dataset.}
  \label{fig:BenthamSamples}
\vspace{-0.5em}
\end{figure}

The ICFHR14 dataset contains text-line images extracted from around
$433$ page images, some examples of which are shown in
Fig.~\ref{fig:BenthamSamples}.
%
It was first used in the ICFHR-2014 HTR competition~\cite{sanchez:2014}
and is now freely available for research purpose at \textsc{zenodo}
(see \ref{ap:software}).
%
%
%

%
This early dataset was carefully prepared by the ICFHR14 organisers so
as to avoid the need of LA and to simplify ``non-essential'' HTR
matters as much as possible. To this end, text lines were manually
detected and extracted and small pieces of text such as marginalia
were ignored.
%
Thus, all the benchmarking results reported so far for this dataset
have been based only on conventional WER, exactly as discussed in
Sec.\,\ref{sec:traditionalWerCer}.  That is, the given training
text-line images and their corresponding GT transcripts were directly
used for model training and the WER was evaluated on the results
achieved for the independent set of test line images.

Here we will use the test-set line images to simulate different
shortcomings typically expected both from HTR and LA.  Main statistics
of this test set are reported later in Table\,\ref{tab:datasets}.
%


\subsection{General Settings to Analyse the Proposed Metrics}
\label{sec:settSimul}

%
%


For each test-set page, the transcripts of the different text-lines
were concatenated into a single word sequence,$^{\ref{foot:tok}}$
following the RO specified in the GT of that page.  From this sequence,
$\WER$, $\bWER$ and $\hWER$ can be computed according to
Eqs.\,(\ref{eq:fullPageWER},\,\ref{eq:werBoW}) and \eqref{eq:hungWER},
respectively.
NSFD, in turn, can be determined according to Eq.\,\eqref{eq:nsfd},
using the alignment derived from the computation of $\hWER$, after the
position renumbering described in Sec.\,\ref{sec:roAssess}.  Finally,
$\CER$ and $\hCER$ can be calculated as explained in
Sec.\,\ref{sec:traditionalWerCer} and at the end of
Sec.\,\ref{sec:hungarian}.


To obtain global values of these metrics for a whole test set
of $K$ page images, let
$\mathcal{C}=\{(X_1,Y_1),(X_2,Y_2),\ldots,(X_K,Y_K)\}$ be the set of
page-level pairs of reference and transcription hypothesis.  We
perform ``micro-averaging'' that somewhat minimises the impact of the
relative page sizes (number of words or characters).  For any metric
$m(\cdot,\cdot)$, the global micro-average, $\bar{m}(\mathcal{C})$,
can be expressed as the weighted sum of values of $m$ computed for
each page:
\vspace{-0.3em}
\begin{equation}\label{eq:mavg}
  \bar{m}(\mathcal{C})~=~
          \frac{\sum_{k=1}^K{N}_{\!k}\,m(X_k,Y_k)}{\sum_{k=1}^K{N}_{\!k}}
\vspace{-0.3em}
\end{equation}
where $m(\cdot,\cdot)$ can be one of the following page-level metrics:
$\WER$, $\bWER$, $\hWER$, $\CER$, $\hCER$ or NSFD.
%
That is, page metric values are weighted by the corresponding number
of reference words (or characters) in the page, ${N}_{\!k}$,
accumulated over all the test-set pages, and finally normalised by the
total number of reference words (or characters for $\CER$).


Among the proposed metrics, only $\hWER$ has a tunable parameter;
namely, the regularisation factor of Eq.\,\eqref{eq:hungRegulCER},
$\gamma$.
%
%
Throughout several tests, it has been consistently found that this
parameter does not require critical tuning.
For one of these typical tests, Table\,\ref{tab:reguTuning} reports
NSFD and $\hWER$ results for increasing values of $\gamma$.
These results were obtained in a controlled RO--alteration experiment,
described in Sec.\,\ref{sec:roRes}, where random swaps were applied to
4 text lines of each image, at distances ranging from 4 to 7 lines apart.

\vspace{-1em}
\begin{table}[htb]
\centering
\caption{ \label{tab:reguTuning}%
  $\hWER$ and NSFD results (in percentage) for increasing values of
  the regularisation factor $\gamma$.  The corresponding $\WER$ and
  $\bWER$ results were $42.6\%$ and $12.4\%$, respectively.
}
\vspace{0.3em}
\begin{tabular}[b]{l|rrrrrrrrrr}
\toprule
$\gamma$
  & 0~~  & 10$^{-4}\!\!$& 0.1 & 1~& 2~&   5~ & 10~  &  20~ &  50~ & 100 \\
\midrule
$\rho$
   & 14.7 & 12.9 & 12.9 & 12.8 & 12.6 & 11.5 &  9.0 &  5.3 &  1.6 &  0.9 \\
$\hWER$
   & 12.4 & 12.4 & 12.4 & 12.4 & 12.5 & 13.8 & 17.9 & 25.4 & 36.5 & 42.7 \\ 
\bottomrule
\end{tabular}
\vspace{-0.3em}
\end{table}

As discussed in Sec.\,\ref{sec:hungarian}, $\rho$ actually decreases
monotonically with $\gamma$, while $\hWER$ is almost constant and
identical to $\bWER$ for a wide range of $\gamma<5$.
According to these and other similar results, the regularization
factor was set to $\gamma=1.0$ for all the experiments presented in
this paper.


\subsection{Inducing Word-level Character Errors,
  While the RO is Kept Essentially Unchanged}
\label{sec:icerRes}

In this experiment we applied increasingly higher character-level
insertions, deletions and substitution distortion to the test-set
reference transcripts, while keeping text lines in their original (correct) RO.
%
%
Two different settings were considered: 1) \emph{``line-level''},
where white-space editing operations are allowed to separate or join
words, and 2) \emph{``word-level''}, where white-space was excluded
from editing operations in order to keep the number of running words
unchanged.

The lowest distortion was chosen so as to induce a CER of $3.25\%$,
which is the CER of real HTR transcripts obtained in a regular
experiment (see Table\,\ref{tab:E2EevalRes}).
Increasing distortion was then progressively applied according to
$\tCER(n)\!=\!3.25n,\,n\!\in\!\{1,2,\ldots,6\}$, until reaching an
induced (or "theoretical'') tCER of $19.5\%$.
%
The distribution of the total tCER into the different character error
types was set proportional to the observed proportions of
substitutions, insertions and deletions.
Further, for line-level distortion, the proportion of white-space
characters was set according to the character error distribution
observed in the real HTR experiment.

\begin{figure}[htb]
  \centering
  \includegraphics[width=.47\textwidth]{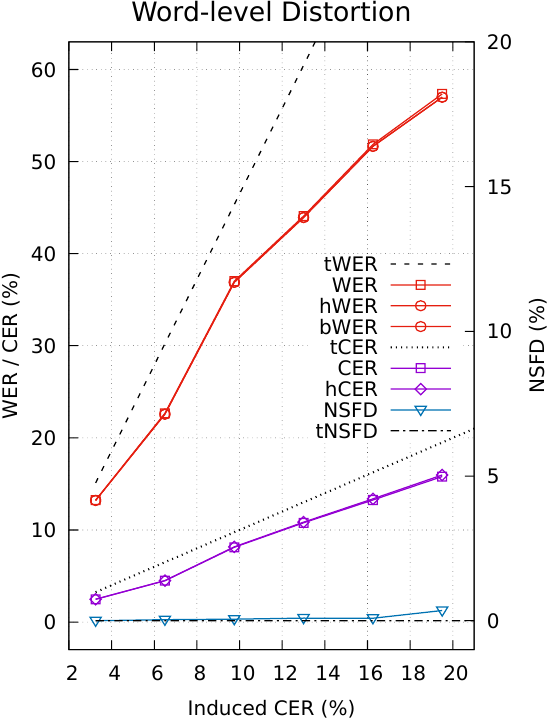}%
  \hspace{2em}%
  \includegraphics[width=.47\textwidth]{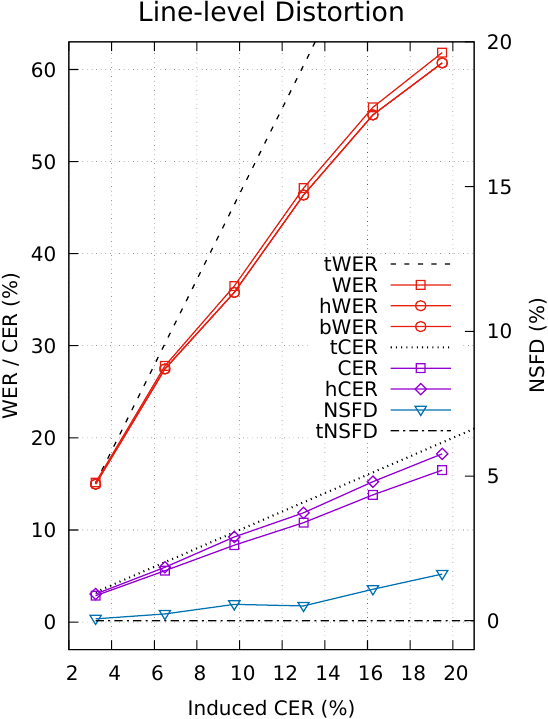}%
  \vspace{-0.8em}
  \caption{Evaluation results for increasingly distorted transcripts,
    as a function of the CER artificially induced by the distortion
    process.
    Left: words are distorted individually avoiding induced white
    space errors to break or join words.
    Right: distortion is applied at full line level, allowing white
    space to be deleted/inserted between/within words.
    Curves with very similar or identical values are depicted with the
    same colour and/or point shape.
    The prefix ``t'' in tCER, tWER and tNSFD indicates the corresponding
    values are theoretically computed.}
  \label{fig:lineDistRes}
\vspace{-0.4em}
\end{figure}

Fig.\,\ref{fig:lineDistRes} plots the empirical $\WER$, $\bWER$ and
$\hWER$ results, along with the theoretical values of induced CER
(tCER, dotted-line) and WER (tWER, dashed-line, calculated according
to $\tWER(n)\!=\!4.65\cdot\tCER(n)$, where $4.65$ is the average word
length in the reference transcripts).
%


\vspace{-0.3em}

Results for the word-level distortion are shown in
Fig.\,\ref{fig:lineDistRes}-left.  As the RO in this case is not
altered at all, the theoretical NSFD (tNSFD) is 0 (horizontal
dash-dotted line).  As expected, all the empirical NSFD values are
also very close to 0.
Moreover, the empirical values of $\WER$, $\bWER$ and $\hWER$ all grow
almost identically for increasing tCER. This also holds for $\CER$ and
$\hCER$.

For line-level distortion the results are shown in
Fig.\,\ref{fig:lineDistRes}-right.  In this case, for large tCER, the
empirical NSFD results become significantly larger than 0, and
$\CER$ is also somewhat larger than $\hCER$.
This is clearly due to the white-space editing operations which, for
large tCER, results in significant variations in the number of
words. The HA need to accommodate these variations by means of
insertions and/or deletions, which explicitly increases the NSFD,
albeit only moderately.


\vspace{-0.3em}
\subsection{Altering Text Line RO for HTR Transcripts with Fixed Word Errors}
\label{sec:roRes}

Here we evaluated the impact of altering the RO of
the real HTR transcripts produced in a regular HTR experiment (namely,
the one whose results are reported in the first row of
Table\,\ref{tab:E2EevalRes} of Sec.\,\ref{sec:e2eRes}).
For the sake of simplicity, alterations considered in this section are
limited to whole-line swapping.  This aims to simulate typical
failures in text-line ordering, often caused by poor (implicit or
explicit) LA of images with multi-column text blocks, marginalia, etc.

For each test-set page with $M$ text lines, the order of line
transcription hypotheses is changed by swapping a given number of line
pairs, $S$, at a given distance or range, $r$.  Line pairs are randomly
selected, but lines already swapped are not candidate for further
swapping.
For a given $S$, depending on the value of $r$, the actual number of
possible swapping on a page may be lower than $S$.
For example, the maximum number of swappable line pairs of a page with
$M=8$ lines, at a distance $r=7$, is only one: the first line with the
last one of that page.

\begin{figure}[htb]
  \centering
  \includegraphics[width=.44\textwidth]{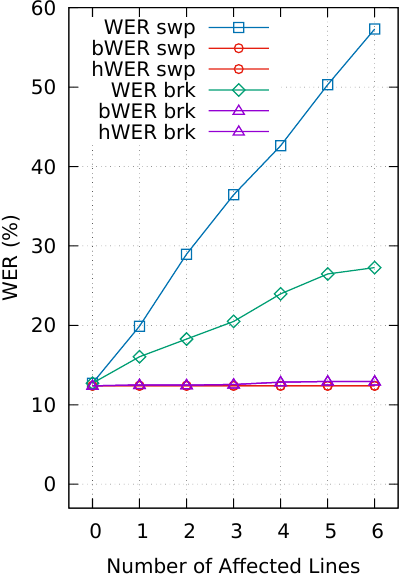}%
  \hspace{2em}%
  \includegraphics[width=.44\textwidth]{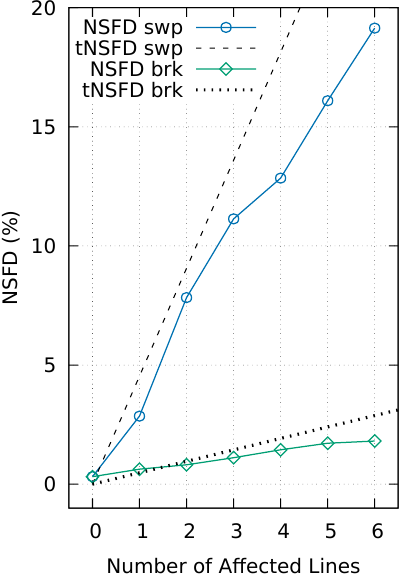}%
  \vspace{-0.3em}
  \caption{Evaluation results on actual HTR transcripts
    where the line order is distorted by random line swaps and
    breaks. Curves with almost identical values are depicted with the
    same colour and symbol. tNSFD corresponds to theoretically computed
    values. The ``swp'' and ``brk'' labels denote line swap and split,
    resp. (see Sec.\,\ref{sec:roRes} and \ref{sec:lnBrkRes}).}
  \label{fig:lineSwapsBreakRes}
\vspace{-0.5em}
\end{figure}

\medskip
For a given range of swap distances $r\in[R',R]$, and a given number
of pages, $K$, the expected NSFD induced by this process,
$\tilde{\rho}$, can be approximated as:
\vspace{-0.3em}
\begin{equation}\label{eq:nsfd-wordLv4}
  \tilde{\rho}(K,S,R',R) ~\approx~
        \frac{S(R'+R)}{K}\,\sum_{k=1}^K\frac{1}{\lfloor M_k^2/2\rfloor}
\vspace{-0.3em}
\end{equation}
where $M_k$ is the number of lines of the $k$-th page.
In our experiments, $K=33,\,R'=4,\,R=7,$ and
$\,\sum_{k=1}^K1/{\lfloor M_k^2/2\rfloor}=0.136$, yielding:
$\tilde{\rho}(S) ~=~ 0.045\,S$. In the right plot of
Fig.\,\ref{fig:lineSwapsBreakRes} shows these expected NSFD values as
the dashed line labelled ``tNSFD swp'').

\medskip
The left plot of Fig.\,\ref{fig:lineSwapsBreakRes} shows WER values
obtained for different (maximum) numbers of swapped lines, where each
value is the average over a range of swap distances $[4,7]$.
As expected, while $\WER$ increases quickly with the number of
swapped lines, the corresponding $\bWER$ and $\hWER$ remain almost
constant.
On the other hand, the right plot shows how the empirical NSFD values
also grow as the number of line swaps increases, more or less closely
following the expected linear tendency (tNSFD~swp).
Fig.\,\ref{fig:lineSwapsBreakRes} also includes WER and NSFD results
of the experiments discussed in the next subsection.




\subsection{Impact of Text Line Splitting Errors}
\label{sec:lnBrkRes}

\begin{figure}[htb]
  \centering
  \fbox{\includegraphics[width=.73\textwidth]{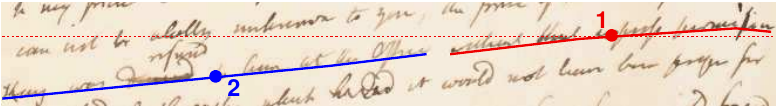}}\\[0.3em]
  \fbox{\includegraphics[width=.85\textwidth]{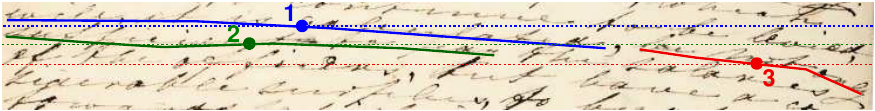}}\\[0.3em]
  \fbox{\includegraphics[width=.70\textwidth]{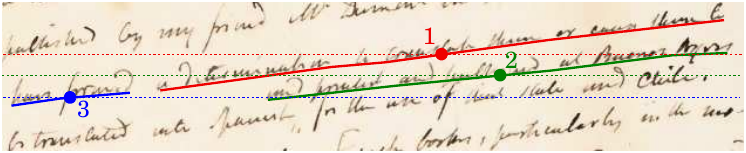}}
  \vspace{-0.3em}
  \caption{\label{fig:splitLines}%
    Real examples from Bentham Papers images 010\_003\_002,
    019\_004\_003, which illustrate text line splitting errors that
    affect RO.  In the simulation experiments these examples
    correspond, top to bottom, to Cases 1, 2 and 3.}
\end{figure}

Finally we check the effect of randomly inserting line-breaks in the
HTR transcripts.  
This aims to simulate (implicit or explicit) line detection errors
which lead to wrong intra-line text ordering.
To this end, the following procedure was carried out for the HTR
transcripts of each test page: 1) $S$ lines are randomly selected. 2)
For each selected line a splitting position is randomly chosen; it can
be at character or word level, with a chance of $1$ to $4$
respectively. 3) The split line fragments are relocated according
to one of these three equiprobable options:
i) the line suffix goes before the prefix, ii) the line suffix goes
after the line next to the selected one, or iii) the line prefix goes
after the line next to the selected one. These cases correspond to
relatively common flaws of (implicit or explicit) LA, which may happen
mainly with highly skewed text images, as illustrated in
Fig.\,\ref{fig:splitLines} (see also Fig.\,\ref{fig:complexExamples}).

\bigskip
As in Sec.\,\ref{sec:roRes}, we can estimate the impact of these RO
alterations on the NSFD metric. Ignoring the effect of word breaks,
the NSFD induced for $K$ page images can be approximated as:
%
\begin{equation}\label{eq:nsfd-wordLv4b}
  \tilde{\rho}(S) ~\approx~
   \frac{7}{6}\frac{S}{K}\sum_{k=1}^K\frac{N_k/ M_k}{\lfloor M_k^2/2\rfloor}
\end{equation}
where, as before, $M_k$ and $N_k$ are respectively the number of lines
and words of the $k$-th page image. For our $K=33$ page images, this
leads to $\tilde{\rho}(S)=0.0048\,S$. The dotted line labelled ``tNSFD
brk'' in the right plot of Fig.\,\ref{fig:lineSwapsBreakRes} shows
these expected NSFD values.

%
Note that, unlike the RO alteration simulation of
Sec.\,\ref{sec:roRes}, here not only the RO is changed (in this case
at a range distance $r=1$), but also some words are distorted because
a line split point may happen to fall within a word, thereby producing
two word fragments.
Such word splits happen with probability $S\,(K/4)/\sum_{k=1}^K\,N_k$
and for each split, two word errors are expected.  In our case, $K=33$
and, for the transcription hypotheses, $\sum_{k=1}^KN_k=6955$.
Therefore, the expected increase of $\bWER$ (and $\hWER$) is
$0.0023\,S$ ($0.23\,S$ in \%), which explains the tiny increase of
$\bWER$-brk and $\hWER$-brk observed in
Fig.\,\ref{fig:lineSwapsBreakRes}.

\subsection{WER--NSFD Correlation and Computational Costs}
\label{sec:concluSimu}


In Sec.\,\ref{sec:icerRes} (Fig.\,\ref{fig:lineDistRes}) we have seen
that, when the amount of character (and word) errors increases without
changing the word order, the NSFD remains essentially constant and
close to 0.  In contrast, all the word and character error metrics
grow almost linearly with the amount of induced character errors.
Moreover, the three word error metrics ($\WER$, $\bWER$ and $\hWER$)
yield almost identical values in all the cases.
On the other hand, we have seen in Sec.\,\ref{sec:roRes}
(Fig.\,\ref{fig:lineSwapsBreakRes}) that if the amount of word errors
is kept constant but the RO of the transcripts is increasingly
perturbed, both $\WER$ and NSFD (and also $\CER$) grow fairly linearly
with the amount of induced RO mismatch.  In contrast, now $\bWER$ (and
$\hWER$) remain practically constant and equal to the value of $\WER$
when HTR and reference transcripts are in the same RO.

\begin{figure}[htb]
  \centering
  \begin{minipage}[t]{.48\textwidth}
    \centering
    \raisebox{0pt}{
      \includegraphics[height=1.33\textwidth]{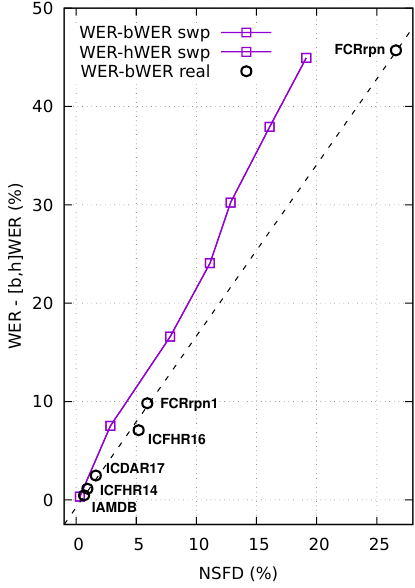}}%
    \vspace{-0.5em}
    \caption{Correlation of $\WER\!-\!\bWER$ (and $\WER\!-$ $\!\hWER$)
      with NSFD. Real results from Table~\ref{tab:E2EevalRes} are
      included, along with a straight line fitted to these points.
      Curves with almost identical values are depicted with the same
      colour and symbol.
      \label{fig:correlation}}
  \end{minipage}%
  \hfill%
  \begin{minipage}[t]{.48\textwidth}
    \centering
    \includegraphics[height=1.33\textwidth]{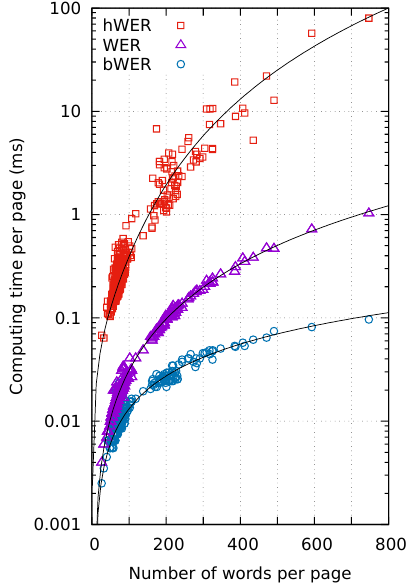}
    \vspace{-0.5em}
    \caption{Computing Times of $\hWER, O(N^3)$, $\WER, O(N^2)$ and
      $\bWER, O(N)$, fitted respectively to polynomials of degrees 3,
      2 and 1 (linear).
      \label{fig:compTimesHung}}
  \end{minipage}
\vspace{-0.5em}
\end{figure}

All these results (those of Fig.\,\ref{fig:lineSwapsBreakRes} in
particular) suggest a strong correlation between NSFD and $\WER$, with
an offset given by $\bWER$ (or $\hWER$). This is explicitly put
forward in Fig.\,\ref{fig:correlation}, where values of
$\Delta\!\WER\eqdef\WER-\bWER$ (and also $\WER-\hWER$) are plotted
against the corresponding values of NSFD.
We also include in this plot a few points corresponding to real
end-to-end evaluation results of some of the experiments that will be
presented in Sec.\,\ref{sec:realEval} (Table\,\ref{tab:E2EevalRes}).
It can be seen that these points also show a fair linear correlation
between $\Delta\!\WER$ and NSFD.


Regarding the relative costs of the different metrics, computing times
are plotted in Fig.\,\ref{fig:compTimesHung} as a function of the
number of words per page.  All the times were measured on the same
computer, using the C++ implementations of $\WER$, $\bWER$, and
$\hWER$ described in \ref{ap:software}.
The points correspond to real end-to-end evaluation of individual pages
and the least-square fitted curves clearly show the different time
complexities of each method.

\vspace{-0.5em}
\section{Examples of Real End-to-End Evaluation}\label{sec:realEval}
\vspace{-0.2em}

The proposed evaluation metrics have been applied to assess end-to-end
HTR systems in real scenarios.  The HTR datasets considered, the
empirical settings and the results obtained are presented in the
following subsections.

\vspace{-0.3em}
\subsection{Datasets and Methods}\label{sec:empiricalSettings}
\vspace{-0.2em}

Besides the historical dataset ICFHR14~\cite{sanchez:2014} already
used in the preceding sections,
four additional datasets were selected to test the proposed evaluation
metrics; namely: the traditional modern handwriting benchmark
IAMDB~\cite{marti2002iam}, and three historical handwriting datasets:
ICFHR16~\cite{sanchez2016} and ICDAR17~\cite{sanchez:2017}, compiled
for the ICFHR-2016 and ICDAR-2017 HTR competitions, and the
\emph{Finnish Court Records} dataset (FCR)~\cite{quiros2022} from the
``Renovated District Court Records'' held by the National Archives of
Finland. Information about how to download each of these datasets is
given in \ref{ap:software}.

IAMDB is a well known modern English handwritten text corpus, gathered
by the FKI-IAM Research Group on the base of the Lancaster-Oslo/Bergen
text Corpus (LOB)~\cite{johansson1978lancaster}. The last released
version (3.0) contains about $1\,500$ scanned text pages,
written by $657$ different writers.

The ICFHR16 dataset encompasses $450$ page images which are 
a subset of the Ratsprotokolle collection, written in old German
and composed of handwritten minutes of council meetings held from 1470
to 1805. One remarkable characteristic of this dataset is that their
text lines are short, each one containing very few (long) words.

The ICDAR17 dataset comprises around $10$K page images,
most of which taken from the \emph{Alfred Escher Letter Collection}.
This collection is mostly written in German, but it also includes
pages in French and Italian. Here, the performance evaluation was
carried out on the pages corresponding to the partition called
``Test-B2'' in~\cite{sanchez:2017}, which was aimed to evaluate not
only text recognition accuracy but also (indirectly) LA performance.

Finally, the FCR dataset consists of $500$ manuscript images
which contain records of deeds, mortgages, traditional lifeannuity,
among others. They were written in Swedish by many hands during the
18th century.  Here, the evaluation was done on $100$ images
($48$ are double-page images) which are a subset of the test partition
used in\,\cite{vidal2021hyp}.
For more details about ICFHR14, ICFHR16, ICDAR17 and FCR datasets,
refer to~\cite{sanchez2019,quiros2022}.

It is important to remark that no tokenization (e.g. to separate
punctuation marks from words) was applied to text references or HTR
transcripts, excepting ICDAR17 whose original references and HTR
results obtained in the associated competition were used.
Table\,\ref{tab:datasets} reports the main statistics
\emph{only for the test sets}, which are the focus of
the proposed evaluation metrics.

\vspace{-0.7em}
\begin{table}[htb]
    \centering
    \caption{
      \label{tab:datasets}%
      Test set main statistics of the evaluated datasets. Except for
      ICDAR17, the running words and lexicon sizes correspond to
      untokenized ``words'', which may include punctuation marks.}
    \vspace{0.3em}
    \small
    \begin{tabular}{l|rrrrrr}
      \toprule[1.5pt]
                      & ICFHR14 &    IAMDB & ICFHR16 & ICDAR17 &      FCR \\
      \midrule
      Number of pages &      33 &      336 &      50 &      57 &      100 \\
      Number of lines &     860 &   2\,915 &  1\,138 &  1\,412 &   6\,183 \\
      Running words   &  6\,966 &  23\,406 &  3\,546 & 14\,460 &  33\,999 \\
      Running chars   & 38\,474 & 123\,090 & 22\,396 & 80\,568 & 214\,785 \\
      Lexicon         &  2\,278 &   6\,398 &  1\,834 &  4\,648 &   9\,890 \\
      Alphabet size   &      82 &       75 &      80 &     104 &       83 \\
    \bottomrule[1.5pt]
    \end{tabular}
\vspace{-0.3em}
\end{table}

%

\vspace{0.7em}
Except ICDAR17 (for which the same transcripts as
in~\cite{sanchez:2017} were used), for each dataset we trained
specific optical character models using the provided training images
and the corresponding reference transcripts.
Character modelling was based on
\emph{Convolutional-Recurrent Neural Networks} (CRNN), trained using
the state-of-the-art freely available PyLaia
Toolkit.\!%
\footnote{\url{https://github.com/jpuigcerver/PyLaia}}
The same setup described in\,\cite{vidal2021hyp} was adopted here to
specify the CRNN topology and meta-parameters.

HTR transcripts of test images were obtained
through two different ways of line extraction:
1) use the line locations and RO given in the GT; and 2) use a
\emph{Region Proposal Network} (RPN)~\cite{lquiros:21}
trained to detect and extract lines with a RO given by their positions
on image, from top-to-bottom and left-to-right as in\,\cite{quiros2022}.
To this end, the same RPN topology and meta-parameter settings
as in~\cite{lquiros:21} was adopted.
%
For both ways of line extraction and each dataset, the corresponding
CRNN model trained with PyLaia was used to decode the extracted line
images. Finally, HTR full-page transcripts were produced by
concatenating the predicted text lines according to their RO
given by the GT or computed by the RPN.

In addition to the above ``classical'' HTR experiments, as an example
of what we consider the ultimate aim of the proposed metrics, we also
test the end-to-end LA\,+\,HTR approach named
\emph{Simple Predict \& Align Network} (SPAN)\,\cite{Coquenet2021}.%
\footnote{\url{https://github.com/FactoDeepLearning/SPAN}}
This model learns to transcribe paragraphs by aligning all the text
line representations via a horizontal feature map unfolding.
%
%
%
By training with the CTC loss strategy, this model learns how to align
input information with the feature map rows and produce a sequential
output, without requiring any specific LA preprocessing.

\vspace{0.5em}
\subsection{Real End-to-End Evaluation Results}\label{sec:e2eRes}
\vspace{0.3em}

Table\,\ref{tab:E2EevalRes} reports performance in terms of the
proposed evaluation metrics for different end-to-end HTR approaches,
tested on the datasets outlined before.  The way of line extraction
and ordering, as well as the HTR system adopted, appear on the columns
labelled LA+RO and HTR, respectively.
Selected values of $\Delta\!\WER=\WER-\hWER$ and NSFD, highlighted
in boldface, are plotted in Fig.\,\ref{fig:correlation}, as already
mentioned in Sec.\,\ref{sec:concluSimu}.

\begin{table}[htbp]
  \centering
  \caption{
    \label{tab:E2EevalRes}%
    Real evaluation results for different datasets, LA and HTR
    approaches.  All values are percentages.  $\Delta\!\WER$ denotes
    $\WER\!-\!\bWER$.  RO mismatch (NSFD) and $\Delta\!\WER$ values
    corresponding to points shown in Fig.\,\ref{fig:correlation} are
    marked in boldface.
    $\WER,\bWER$ and $\hWER$ $95\%$ confidence intervals are narrower
    than $\pm$1.6\% for ICFHR16 and $\pm$1\% for all the other
    datasets. The LA+RO approach ``TRB'' for ICDAR17 stands for
    Transkribus platform.}
  \vspace{0.5em}
  \small
  \extrarowheight=-0.2pt 
  \begin{tabular}{l|ll||rr|rrr|rr}
    \toprule[1.5pt]
    \raisebox{-4pt}{Dataset}\raisebox{5pt}{\!\!\!\!\!\!Metric\!\!}
    & LA+RO & HTR & NSFD &\!\!\!$\Delta\!\WER$ & $\WER$ & $\!\!\!\bWER$
                            &   $\!\!\!\hWER$ & $\CER$ & $\!\!\!\hCER$ \\
    \midrule[1.5pt]
    \multirow{2}{*}{ICFHR14}
    & GT  & CRNN & 0.3 & 0.3 & 12.7 & 12.4 & 12.4 & 3.3  & 4.0 \\
    & RPN & CRNN & {\bf 0.9}& {\bf 1.1}& 17.4 & 16.3 & 16.3 & 5.5  & 5.9 \\
    \midrule
    \multirow{3.5}{*}{IAMDB}
    & GT  & CRNN & 0.6 & 0.5 & 27.0 & 26.5 & 26.5 & 7.5  & 8.2 \\
    & RPN & CRNN & {\bf 0.7} & {\bf 0.5} & 27.8 & 27.3 & 27.3 & 7.9  & 8.7 \\
    \cmidrule{2-10}
 &\multicolumn{2}{c||}{SPAN}& 0.5 & 0.6 & 26.7 & 25.9 & 26.0 & 7.5  & 8.3 \\
    \midrule
    \multirow{3.5}{*}{ICFHR16}     
    & GT  & CRNN & 0.3 & 0.6 & 27.7 & 27.1 & 27.2 &  5.7 & 6.6 \\
    & RPN & CRNN & {\bf 5.2}& {\bf 7.1} & 33.5 & 26.4 & 26.6 & 13.7 & 6.5 \\
    \cmidrule{2-10}
    &\multicolumn{2}{c||}{SPAN}& 1.3 & 1.6 & 31.5 & 29.9 & 30.0 & 10.7 & 10.9\\
    \midrule
    \multirow{2}{*}{ICDAR17}
    & GT  & CRNN & 1.4 & 2.2 & 18.6 & 16.4 & 16.5 &  6.3 & 6.6 \\
    & TRB & CRNN & {\bf 1.6} & {\bf 2.5} & 20.1 & 17.6 & 17.7 &  7.0 & 7.1 \\
    \midrule
    \multirow{3}{*}{FCR}
    & GT  & CRNN & 0.8 & 1.1 & 25.2 & 24.1 & 24.4 &  5.6 & 6.4 \\
    & RPN & CRNN &{\bf 26.6}&{\bf 45.7}& 72.4 & 26.7 & 27.0 & 50.8 & 8.5 \\
    & RPN1 & CRNN & {\bf 5.9} & {\bf 9.8} & 36.5 & 26.7 & 27.0 & 15.1 & 8.2 \\
    \bottomrule[1.5pt]
  \end{tabular}
  \vspace{-0.2em}
\end{table}

In all the cases, $\hWER$ is slightly higher than or identical to
$\bWER$ and both are always smaller than $\WER$, as discussed in
Secs.\,\ref{sec:bow},\ref{sec:hungarian} (and summarises in
Sec.\,\ref{sec:summMetrics}) -- and as expected from the simulation
results of Sec.\,\ref{sec:simuExp}.
Also as expected, all the HTR approaches which use the (perfect) text
lines and RO given by the GT, achieve lower NSFD and $\Delta\!\WER$,
compared with other approaches involving automatic line detection.

Another important general remark is the fairly tight correlation
observed between NSFD and $\Delta\!\WER$.  It is more clearly seen for
results more or less affected by RO issues, specifically those
highlighted in boldface which, as commented, are also plotted in
Fig.\,\ref{fig:correlation}.  This further endorses the discussion in
Sec.\,\ref{sec:concluSimu} and adds empirical support to consider
$\Delta\!\WER$ as a suitable metric to put forward LA or, in general,
RO problems.

%

IAMDB has a very simple RO structure and no significant differences
exist among the different error rate metrics.  To a lesser extent, the
same can be said for ICFHR14.

The case of ICFHR16 is worth commenting.  The $\WER$ achieved by
RPN--CRNN ($33.5\%$) is significantly higher than the $\bWER$
($26.4\%$), leading to $\Delta\!\WER=7.1\%$.  This makes it clear that
the RO provided by RPN LA is far from perfect, an issue directly
supported by the fairly high value of NSFD ($5.20\%$).  As discussed
later in more detail, most RO errors are due to marginalia transcripts
for which the system fails to place in the correct RO.
%
%

Also interesting is the case of FCR, which contains a mixture of
single- and double-page images.  For double-page images, the regular
RPN settings (denoted in the table just as ``RPN'') dramatically fail
to separate the lines of each page and render them in the correct RO.
So, even though the individual words are fairly well recognised (with
$\bWER=26.7\%$), the conventional $\WER$ is exorbitant ($72.4\%$).
This leads to a very large $\Delta\!\WER$ ($45.7\%$) which clearly
shows the massive RO mismatch, also reflected by the very large value
of NSFD ($26.6\%$).
Of course, this experiment was only aimed at providing a clear
illustration of the behaviour of proposed metrics.  So we also tested a
more reasonable LA approach (called ``RPN1'' in the table).  In this
approach, when a double-page is identified, each detected text line is
classified as belonging to the left or to the right page and then the
usual RPN RO is applied page-wise.
This approach provides identical individual word recognition
performance ($\bWER=26.7\%$) and greatly solves the RO issues --
albeit not completely, as assessed by the still high values
$\Delta\!\WER=9.8\%$ and NSFD$=5.92\%$.
%

Regarding $\CER$ and $\hCER$ results, in general they reflect similar
tendencies as $\WER$ and $\hWER$ when RO issues are involved.  Note
however that, as discussed in Sec.\,\ref{sec:hungarian}, $\hCER$ is
only an approximation and is not as directly and faithfully comparable
with $\CER$ as $\hWER$ is with $\WER$.
%

%
The SPAN (true full-page) approach, was tested on two datasets.
Results for IAMDB are comparatively good in terms of word and
character error metrics and also in terms of RO as assessed by NSFD
and $\Delta\!\WER$.

The SPAN results for ICFHR16 deserve a more detailed analysis.  The
reading-order independent word recognition results
($\hWER\!\approx\!\bWER\!=\!29.9\%$) are sensibly worse than those of
RPN+CRNN discussed above ($\hWER\!\approx\!\bWER\!=\!26.4\%$), while
the conventional WER is somewhat better ($31.5\%$ vs. $33.5\%$).  So
the $\Delta\!\WER$ for SPAN is significantly lower ($1.6\%$
vs. $7.1\%$) -- which is also consistent with NSFD ($1.3\%$
vs. $5.2\%$).  This indicates that the transcripts provided by SPAN
have more word errors but are in significantly better RO than
RPN+CRNN.

\medskip
To better understand these results, we can gather additional
evaluation clues from the distribution of $\bWER$ errors.  In this
case, from $\bWER\!=\!29.9\%$, $25.7\%$ errors are substitutions,
$0.3\%$ insertions and $3.9\%$ deletions.  So we observe that SPAN
makes many word deletions, around $10$ times more than RPN+CRNN (with
$0.4\%$ deletions, $0.6\%$ insertions and $25.4\%$ substitutions).
%
%
%
A closer look at the SPAN transcripts reveals that, indeed, SPAN
almost systematically delete (i.e., fails to detect and recognise) the
many marginalia lines existing in the ICFHR16 images.  Clearly, while
the RO is hardly affected by this fact, there is a noticeable impact
on the reading-order independent recognition accuracy, evidenced by
the relatively larger values of $\bWER$ and $\hWER$.

As a specific example of this general fact,
Fig.\,\ref{fig:evalExample} shows an ICFHR16 page image, along with
its GT, RPN+CRNN and SPAN transcripts.
In the RPN+CRNN transcript, the lines corresponding to the marginal
note (in red) are correctly detected and all their words recognised
(with two errors).  However, they are mixed with the lines of the last
paragraph (in blue).  The total number of word errors is $22$
($31.4\%$), but because of the mixed marginal lines, the RO is rather
poor, as properly reflected by $\Delta\!\WER\!=\!10\%$.
For the SPAN transcript, $\hWER\!=\!\bWER\!=\!31.4$, exactly the same
as for RPN+CRNN.  But, as suspected, it has completely failed to
transcribe the marginal note words.  However, all the transcribed
words are in good RO, a fact faithfully reflected by
$\Delta\!\WER\!=\!1.4\%$.

\definecolor{lightgray}{rgb}{0.85, 0.85, 0.85}
\begin{figure}[htb]
\vspace{-0.3em}
  \centering
  \setlength{\fboxsep}{1pt}
  \def\s{-.5ex}
  \def\ss{-.2ex}
  \raisebox{-1pt}{
  \includegraphics[width=.40\textwidth]{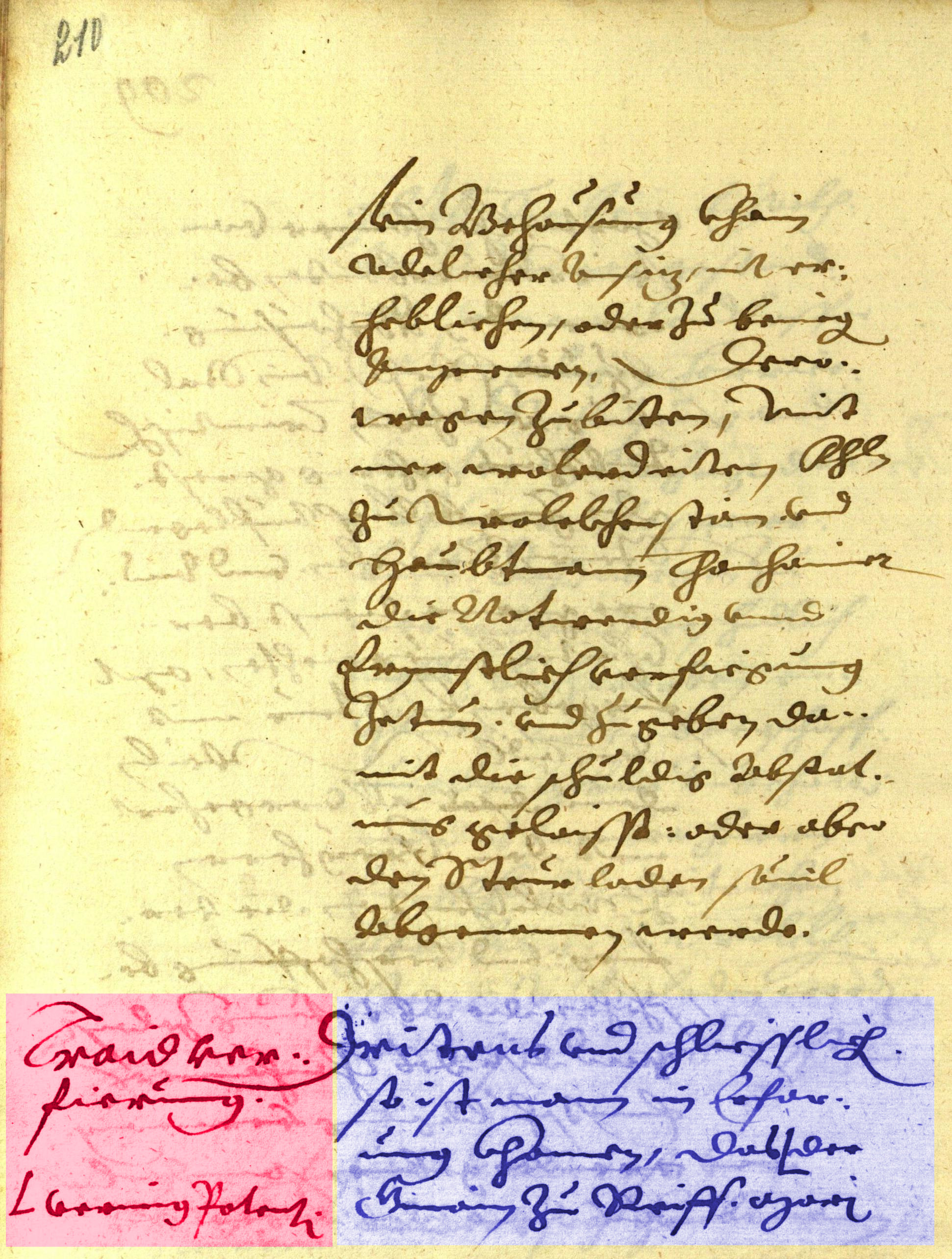}}%
  \fbox{
  \scalebox{.912}{
  \begin{minipage}[b]{.28\textwidth}
    210\\[\s]
    sein Behaūsūng khain\\[\s]
    Adelicher Ansiz, nit er¬\\[\s]
    heblichen, oder Zū benieg\\[\s]
    Angenomen. Dero¬\\[\s]
    wegen Zū biten, Nit\\[\s]
    mer wolerdeiten Fhn\\[\s]
    Zū Wolckhenstain vnd\\[\s]
    Haūbtmann Thanhaimer\\[\s]
    die Notwendig vnnd\\[\s]
    Ernnstlich verfiegūng\\[\s]
    Zetūen. vnd Zū geben da¬\\[\s]
    mit die schūldig Abstat¬\\[\s]
    ūng gelaisst: oder aber\\[\s]
    den Steūr laden sōuil\\[\s]
    Abgenomen werde.\\[\s]
    \blue{Dritens vnd schliesslich.}\\[\s]
    \blue{so ist mann in Erfar¬}\\[\s]
    \blue{ūng khomen, das der}\\[\s]
    \blue{Gmain Zū Keifs,.Mori}\\[\s]
    \red{\bf Traid ver¬}\\[\s]
    \red{\bf fierūng.}\\[\s]
    \red{\bf vermieg Patent.\\[-2ex]}
  \end{minipage}}}%
  \fbox{
  \scalebox{.911}{
  \begin{minipage}[b]{.28\textwidth}
    210\\[\s]
    sein Behaūsūng khain\\[\s]
    \colorbox{lightgray}{Oodelicher} Ansiz, nit er¬\\[\s]
    heblichen, oder Zū \colorbox{lightgray}{benag}\\[\s]
    \colorbox{lightgray}{Anzenemen}, Dero¬\\[\s]
    wegen Zū biten, \colorbox{lightgray}{Vit}\\[\s]
    mer \colorbox{lightgray}{wolerd} \colorbox{lightgray}{eitem} \colorbox{lightgray}{Khln}\\[\s]
    Zū \colorbox{lightgray}{Grolckherstain.} vnd\\[\s]
    Haūbtmann \colorbox{lightgray}{Thonhaimer}\\[\s]
    die Notwendig vnnd\\[\s]
    Ernnstlich verfiegūng\\[\s]
    \colorbox{lightgray}{Zetūn.} vnd Zū geben da¬\\[\s]
    mit die schūldig \colorbox{lightgray}{Abstat.}\\[\s]
    \colorbox{lightgray}{nūng} gelaisst: oder aber\\[\s]
    den Steūr laden \colorbox{lightgray}{soūil}\\[\s]
    \colorbox{lightgray}{Abgemomen} werde.\\[\s]
    \blue{\colorbox{lightgray}{Gritens} vnd \colorbox{lightgray}{schliesslich}}\\[\s]
    \red{\bf Traid ver¬}\\[\s]
    \red{\bf fierūng.}\\[\s]
    \blue{so ist mann \colorbox{lightgray}{im} \colorbox{lightgray}{Erfar}
    }\\[\s]
    \blue{ūng khomen, das der}\\[\s]
    \red{\bf \colorbox{lightgray}{vermig} \colorbox{lightgray}{Potent}}\\[\s]
    \blue{Gmain Zū \colorbox{lightgray}{Rriff.} \colorbox{lightgray}{Marj}\\[-2ex]}
  \end{minipage}}}\\[.5ex]
    \fbox{
  \scalebox{.94}{
  \begin{minipage}[b]{\textwidth}
  \colorbox{lightgray}{10} sein Behaūsūng khain Adelicher Ansiz, \colorbox{lightgray}{niter¬} heblichen, oder Zū benieg \colorbox{lightgray}{Anzenemen,} Dero¬ wegen Zū biten, \colorbox{lightgray}{Nint} mer \colorbox{lightgray}{wolerd} \colorbox{lightgray}{eitem} \colorbox{lightgray}{khln} Zū \colorbox{lightgray}{irolckherstain.} vnd \colorbox{lightgray}{Haūbtmannnhai} \colorbox{lightgray}{Thomer} die Notwendig vnnd \colorbox{lightgray}{Ernnstliich} verfiegūng \colorbox{lightgray}{Zetūm.} vnd Zū geben da¬ mit die schūldig Abstat¬ ūng \colorbox{lightgray}{gelaisso:} oder aber den Steūr laden \colorbox{lightgray}{soūil} \colorbox{lightgray}{Abgemamen} werde. \blue{Dritens vnd schliesslich. so ist mann in Erfar¬ ūng khomen, das der \colorbox{lightgray}{Gmaim} Zū \colorbox{lightgray}{Reiff.} \colorbox{lightgray}{Marj} \colorbox{lightgray}{.}}
  \end{minipage}}}\\[-0.3em]

\caption{ \label{fig:evalExample}%
  A page from the ICHFR16 dataset (ID: Seite0418).
  The red and blue texts and shadings correspond to the text blocks
  affected by RO issues, while word recognition errors are marked
  with shadowed boxes.  The top-middle panel is the reference transcript.
  The right-panel shows the RPN-CRNN's hypothesis, with
  $\bWER=\hWER=31.4\%$ and $\Delta\!\WER=10\%$
  which fairly reflects the RO errors caused by poor LA.
  The bottom panel shows the SPAN's hypothesis, which clearly failed
  to detect and recognise all the marginal note words (in red
  colour).  As compared with RPN-CRNN, SPAN has produced the same
  amount of word errors ($\bWER=\hWER=31.4\%$), but the transcript is
  in better RO, a fact fairly reflected by $\Delta\!\WER=1.4\%$.
}
\end{figure}

\section{Related Works}\label{sec:related}
\vspace{-0.3em}

The problem of assessing the quality of full-page automatic
transcripts, taking into account LA and/or RO errors has been
addressed in many previous works.  In this section, we briefly review
the literature both in HTR and other research fields where alternative
metrics have been proposed in this regard.  The review is organised
into four topics, corresponding to what we consider the four main
contributions of our work.

\vspace{-0.2em}
\subsection{LA and/or RO Awareness in HTR Metrics}

Almost all the works cited in this section consider only
\emph{printed} (historical) documents and the task of full-page,
end-to-end evaluation is always more or less explicitly linked to
(geometric) issues caused by faulty LA---see eg.~\cite{naoum2019}
for a recent work in this category.
Going deeper in this direction, Antonacopoulos, Clausner and
Pletschacher are among the earlier authors who explicitly put forward
the importance of this problem and its relation with RO difficulties ---
and propose pioneering practical approaches for RO-aware
evaluation~\cite{clausner2013,Pletschacher2015}.
%

Note, however, that with the exception
of~\cite{clausner2013,Pletschacher2015} (and others discussed in the
coming subsections), these works are \emph{not} directly concerned
with transcription \emph{evaluation}.  Some produce end-to-end
transcription results, while others deal with LA and/or RO methods;
but all need evaluation metrics and some of them make proposals that
seem adequate to assess their results.
On the other hand, as far we can tell, none of the cited works makes
a convincing assessment of the adequateness of the proposed metrics
for general-purpose evaluation and benchmarking of full-page
transcripts of text images.

In comparison, our work explicitly analyses and proposes
general-purpose metrics which are agnostic of geometry and other
details of LA.  We also report comprehensive results that support the
adequateness of these metrics for unbiased evaluation of the overall
quality of end-to-end transcription results of handwritten (or
printed) text images.

\vspace{-0.2em}
\subsection{Metrics Related With the Bag of Words}

BoW-based assessment appears in~\cite{Pletschacher2015}
and~\cite{Strobel2020}, and it has been used in several ICDAR
competitions~\cite{%
Clausner2017,clausner2017a,ClausnerAP19}.
In these papers and competitions the Bag of Words concept was used to
define evaluation measures based on, or related with recall (or missed
words) and precision (or falsely detected words), generally combined
into a kind of F-measure referred to
``success rate''~\cite{antonacopoulos2017}.
However, the formal details of these measures are not sufficiently
documented and most probably they are largely unrelated with the
metrics we are proposing in this paper.
%
%
%
Moreover, by relying on misses and false detections, the ``success rate''
implicitly overlooks word substitutions, thereby making it difficult to
establish meaningful relations with the traditional WER.

The definition of $\bWER$ in Eq.\,\eqref{eq:werBoW}
explicitly considers substitutions, thereby making
it almost identical to $\hWER$ and allowing for a proper comparison
with the $\WER$.  This leads to the introduction of $\Delta\!\WER$,
which proves to be a very convenient way to measure RO logical
mismatch.

It is worth mentioning that our definition of $\bWER$ is not new.  The
idea was first suggested in~\cite{ney2000} to obtain a rough measure
of the quality of Machine Translation (MT) results disregarding word
order.  Under the name ``Position-independent Error Rate'' (PER),
that idea was later presented more formally
in~\cite{popovic2011}.  By looking closely at the proposed
formulation, one can observe that the core computation is indeed
essentially the same as that of our Eq.\,\eqref{eq:werBoW}.

\vspace{-0.2em}
\subsection{Metrics Related With the Hungarian Algorithm}\label{sec:relatedHA}

The HA has been adopted in many document analysis and recognition
tasks, many of them related with full-page, end-to-end training and/or
text image recognition~\cite{tensmeyer2019training,long2022towards}.
It has been proposed as well for other miscellaneous tasks such as
invoice analysis~\cite{palm2017cloudscan}, pairing different versions
of historical manuscripts~\cite{kassis2017alignment}, and reassembling
shredded document stripes~\cite{liang2019eassembling}, to name a few.
All these works are completely unrelated with evaluation of HTR
transcripts, which is the topic of this paper.

Among the works which explicitly deal with evaluation, we should
mention an interesting early work in the field of Computer Vision,
which considers the evaluation of visual objects
detection~\cite{liu2002optimal}. Several works on LA
make use of the HA to evaluate results of \emph{line detection} and/or
\emph{text region segmentation}~%
\cite{yin2009linesegeval,garz2013linesegeval}.
While this task may seem similar to ours, the overall framework is
quite different.  In these proposals, the elements to be paired are
image regions, and the pairing criterion is strongly based on region
geometry information.
In contrast, our proposal is applied to transcripts, represented just
by character strings.  And evaluation is completely blind to the
existence of text lines and explicitly ignores geometric features of
the text images and/or their GT annotations.

It is worth mentioning that our point of view in this matter is
similar to the one adopted in~\cite{soundararajan2006evaluation} for
assessment of video OCR results.  However, the metric proposed
in~\cite{soundararajan2006evaluation} aims to assess not only the
quality of the transcripts (and their RO), but also the positions of
the detected and recognised words in the image.  Therefore, this
evaluation approach mixes geometric and text criteria, which is
contrary to the principles adopted in our work.

Perhaps the most interesting proposal that is close to our work is the
so called ``Flexible Character Accuracy'' metric~\cite{ClausnerPA20}
(FCA).  It is based on computing the character edit distance between
two chunks of text by iteratively comparing the lines with minimum
edit distance, following a greedy strategy.
%
%
%
The method is further based on several heuristics which need four
weighting factors
to control how much relevance is given to the offset and length
difference of the matched strings.  Additionally, unmatched substrings
are considered insertion or deletion operations, so they are added
as a penalty to the whole result.
This metric was used to assess HTR transcription results in the ICDAR
2019 competition on Recognition of Documents with Complex
Layouts~\cite{ClausnerAP19}.

In our opinion, FCA does succeed in providing a reasonable word
accuracy score which is fairly RO-independent. Nevertheless, it has
two important drawbacks.
First, it is just based on a greedy, suboptimal solution to a line
matching or assignment problem, for which the here proposed
regularised HA would provide an optimal solution.
%
%
In comparison, the approaches here proposed ensure optimal word
pairings and, moreover, they do not need to assume any kind of
LA units such as text blocks or lines.
%
Second, FCA heavily depends on several tunable weights.  Indeed, in
the experiments, the reported results correspond to
a best-scoring combination of parameters for each algorithm run.
Clearly, this makes the method too dependent on the datasets
considered, which would become problematic for general-purpose
benchmarking of full-page transcription results.


\medskip
In addition to the above discussions, perhaps our most important
contribution to the use of the HA for HTR evaluation is to introduce a
\emph{regularised} HA version.
Thanks to the proposed regularisation term, the HA not only minimises
the character edit distance between the paired words, but also avoids
as far as possible word order mismatch, as measured by the NSFD.

Such an enhancement has allowed us to define a HA WER (hWER) which
exhibits all the desired properties:
a) it yields essentially the same results as the bag-of-words WER
(bWER) and thereby provides a proper RO-independent evaluation of
individual word recognition performance;
b) it provides practically the same results as the conventional WER,
whenever reference and system transcripts are in the same RO;
and c) it produces the alignments needed to compute a RO-independent
character error rate and used by NSFD to explicitly measure RO
mismatch.

\vspace{-0.3em}
\subsection{Integrating Evaluation of WER and Reading Order Mismatch}
\label{sec:mixROandWER}
\vspace{-0.1em}

All the works dealing with full-page, end-to-end HTR need to assess not
only word recognition performance, but also the impact of errors due
to flaws in (explicit or implicit)
LA~\cite{Bluche2016,wigington2018start,Coquenet2021}. %
%
Of course, the main focus in these works is on the proposed HTR
methods; so they do not generally pay much attention to how to
properly measure the performance they achieve.

A popular idea is to measure LA errors using conventional LA metrics
and then make do with conventional WER or CER to measure word or
character recognition errors.  Finally, both measures are somehow
combined to obtain a single scalar figure which hopefully represents an
``overall performance'' metric~\cite{coquenet2023dan}.
In a similar vein, but explicitly devoted to HTR evaluation, the work
presented in~\cite{LeifertLGL19},goes deeper in the metric combination
idea, with daunting mathematical formulation.  However, this is a
utterly theoretical work which does not provide any empirical evidence
that would support the proposed formulation or methods in practice.
%

As we see it, the metric combination idea has several drawbacks: 1) as
discussed throughout this paper, if reference and system transcripts
are \emph{not} in the same RO, conventional WER or CER systematically
provide misleading word recognition performance values -- and any
combination of misleading values is obviously also misleading; 2)
metric combination requires adequately tuned weights which are
impossible to adjust for general-purpose benchmarking; and 3) the
required GT is expensive because of the effort entailed by manual
annotation of LA geometric details.

Another idea that has been adopted in some
works~\cite{sanchez:2017,wigington2018start,tensmeyer2019training} is
to assess the overall quality of system transcripts using the so
called ``BLEU'' measure~\cite{papineni2002bleu}.  It is borrowed from
the field of Machine Translation and is based on matching n-gram
frequencies of the system transcript with those of the GT reference.
While this idea avoids the complications and exceedingly high cost of
taking into account LA geometric details, it does suffer form the same
problems of directly using the conventional WER; namely, it jumbles
errors coming from different flaws and it often fails to provide the
kind of insights needed for system improvement.

In contrast with the methods discussed above, the evaluation framework
proposed, developed and assessed in this paper, favours a two-fold
evaluation approach which completely decouples intrinsic word
recognition errors from RO errors caused by poor (explicit or
implicit) LA.

Before closing this section it is worth to cite the work presented
in~\cite{Strobel0CSHS22}, which aims at assessing HTR results without
resorting to GT reference annotations.  While this is indeed an
interesting prospect, it is completely unrelated with the aims and
methods discussed in this paper.

\vspace{-0.2em}
\section{Concluding Remarks}\label{sec:concl}
\vspace{-0.2em}

In classical HTR experiments each relevant text-line image is given and
accuracy is adequately assessed using conventional WER and CER.
When moving to an end-to-end full-page transcription scenario,
page-level accuracy is often being assessed using two very different
metrics: geometric accuracy of layout analysis \emph{and} WER/CER.
We consider that this assessment approach is doubly misleading.
First, geometric accuracy seldom matches well with logical relation
between relevant image elements (text lines).  Second, WER values are
systematically tainted with false word recognition errors caused by
well recognised words which are not placed in in the ``correct'' order.

We argue that methods which aim at end-to-end processing, or at full
integration of layout analysis with word recognition at page-image
level, need assessment criteria which do \emph{not} rely on any kind
of geometric accuracy.
Having this in mind, we have proposed page-level assessment approaches
which: a) are geometry agnostic, b) provide a measure of word
recognition accuracy which does not depend on word reading order, and
c) provides a measure of logical mismatch of transcription elements
(words or lines) which is largely independent on the accuracy with
which individual words are recognised.

As a basic, simple and computationally cheap method to assess word
recognition errors with independence of reading-order, we advocate for
a \emph{reformulated} version of the popular bag-of-words WER, which
we refer to as bWER.
It should be pointed out, however, that the bWER does have some
applicability limitations.  Specifically, as commented in
Sec.\,\ref{sec:bow} and illustrated in Example 3a (\ref{ex:3}), it can
provide optimistically low values if the evaluated transcripts have
many word repetitions.  Clearly, the probability that a chunk of text
contains repeated words grows with the size of the text.
Consequently, bWER is prone to become increasingly optimistic as the
size of the evaluation sample (e.g., page image transcript) becomes
larger.
This is thoroughly studied in~\cite{toselli23} and the results show
that, in general, bWER can be safely used for typical page sizes and
text densities, up to a some hundreds words per page, or even much
larger in some datasets.

In addition, we have introduced another reading-order independent WER,
called hWER, which is based on a \emph{new}, \emph{regularised}
version of the Hungarian Algorithm.
Both bWER and hWER provide almost identical results, but hWER is much
more computationally expensive.  However, the proposed regularised
Hungarian Algorithm underlying the hWER also produces word alignments
which can be used to compute specific reading-order metrics such as
the Normalised Spearman Footrule Distance (NSFD).  Moreover, if system
and references transcripts are in the same reading order, these
alignments very closely approach the traditional word-to-word
sequential ``traces'' underlying the word edit distance assumed in the
classical WER.

The proposed methods are analysed both formally and with the help of
illustrative examples, as well as through a series of partially
simulated experiments.
Finally we have applied state-of-the-art line detection and HTR
methods to a good number of popular benchmark tasks and assessed the
achieved end-to-end accuracy using the proposed metrics.

An important conclusion from both simulated and real assessment
experiments is that the bWER is ideal in practice to assess the
performance of recognising individual words, with full independence of
how these words are ordered in the reference transcripts or in the HTR
transcription hypotheses.  Moreover, empirical evidence also shows
that bWER is almost identical to the classical WER in the classical,
simplified HTR experimental setting where the same reading order for
reference and system transcripts is (rather artificially) guaranteed.

Another important conclusion is that the difference between WER and
bWER ($\Delta\!\WER$) is a very good indicator of the amount of logical
or reading-order mismatch between reference and system transcripts.
Our experiments show that this difference graciously correlates,
almost linearly, with the NSFD, which explicitly measures the
reading-order mismatch.
Thanks to this correlation, the NSFD (which is rather complex and
requires alignments yield by the expensive Hungarian Algorithm)
becomes largely unnecessary.  So, both the individual word recognition
accuracy \emph{and} the degree of logical or reading order mismatch
between (page-level) transcripts, can be assessed using just the
well-known WER and (the properly redefined version of) bWER.

\medskip
\emph{Therefore, our closing recommendation for benchmarking end-to-end
  full-page transcription systems is to provide these two assessment
  figures:} $\bWER$ and $\Delta\!\WER \,\eqdef\, \WER\!-\!\bWER$.\!\!
\medskip

Although both WER and bWER are simple and well known, in
\ref{ap:software} we provide publicly available software to reliably
compute these two metrics, along with the other auxiliary metrics we
have used in this work, the Regularised Hungarian Algorithm WER (hWER)
and the Normalised Spearman Footrule Distance (NSFD), based on the
hWER.%

Following the concepts and results here presented, in future works we
aim to develop adequate loss functions that allow training end-to-end
HTR systems which explicitly optimise the here proposed assessment
criteria.


\section*{Acknowledgements}
\vspace{-0.5em}

This paper is part of the I+D+i projects: PID2020-118447RA-I00
(MultiScore) and PID2020-116813RB-I00a (SimancasSearch), funded by
MCIN/AEI/10.13039/501100011033.
The first author research was developed in part with the Valencian
Graduate School and Research Network of Artificial Intelligence
(valgrAI, co-funded by Generalitat Valenciana and the European Union).
The second author is supported by a
Mar\'{i}a Zambrano grant from the Spanish Ministerio de Universidades
and the European Union NextGenerationEU/PRTR. The third author is
supported by grant ACIF/2021/356 from the ``Programa I+D+i de la
Generalitat Valenciana''.

%
\vspace{-0.5em}
\bibliographystyle{plain}
{\small
 \bibliography{htrPageEval.bib}
\bigskip\bigskip\bigskip\bigskip
}

\vspace{-1.5em}
\appendix

\section{Examples}\label{ap:ex}

\subsection{Example~1}\label{ex:1}

Computation of the edit distance $\edit(x,y)$ and the corresponding
trace $\mathcal{T}(x,y)$.
Deleted and inserted words are marked with red and blue colour, respectively.
See also footnote \ref{foot:tok}.
%
\vspace{-0.8em}
\begin{figure}[htb]\centering
~~~\includegraphics[height=.191\linewidth]{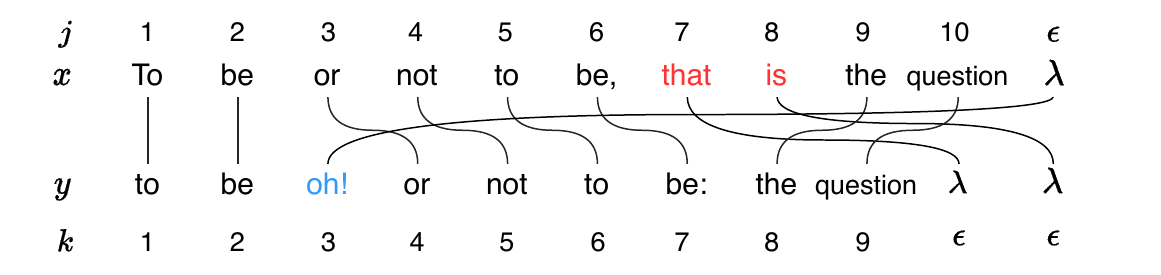}%
\vspace{-1.5em}
\end{figure}
%
\begin{eqnarray*}\label{eq:exED}
  \mathcal{T}(x,y)\!\!
    &=& (1,1),(2,2),(\epsilon,3),(3,4),(4,5),(5,6),(6,7),
        (7,\epsilon),(8,\epsilon),(9,8),(10,9)~~~~~\\
  \edit(x,y)\!\!
    &=& 1+0+1+0+0+0+1+1+1+0+0 ~=~ 5 ~~~~~ (i=1, s=2, d=2)
\end{eqnarray*}

\vspace{0em}
\subsection{Example~2}\label{ex:2}

Computation of the NSFD for a given alignment $\mathcal{A}'$.
%
The original word positions are denoted by $j',k'$, while $j,k$ reflect
the renumbering applied to circumvent the indirect effects of
deletions and insertions.  This converts the original alignment
$\mathcal{A}'(X,Y)$ into $\mathcal{A}(X.Y)$, used in
Eq.\,\eqref{eq:nsfd} to compute the NSFD.
To account for the unit-cost contribution of insertions and deletions,
it is assumed that $|-,\epsilon|=|\epsilon,-|=1$.
\vspace{-0.8em}
\begin{figure}[htb]\centering
\includegraphics[height=.29\linewidth]{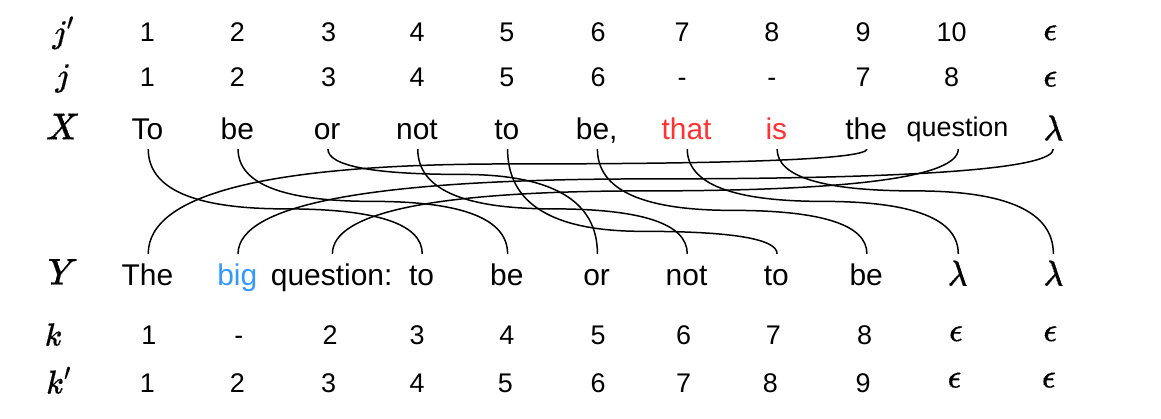}%
\vspace{-1.4em}
\end{figure}
\begin{eqnarray*}\label{eq:exNSFD}
  \mathcal{A}'(X,Y)\!\!
      &=& (1,4),(2,5),(3,6),(4,7),(5,8),(6,9),
          (7,\epsilon),(8,\epsilon),(9,1),(10,3),(\epsilon,2)~~~~~\\[-0.3em]
  \mathcal{A}(X,Y)\!\!
      &=& (1,3),(2,4),(3,5),(4,6),(5,7),(6,8),(\text{-}\,,\epsilon),
          (\text{-\,},\epsilon),(7,1),(8,2),(\epsilon,\,\text{-})~~~~~\\
  \rho(X,Y)\!\!
      &=& (2+2+2+2+2+2+1+1+6+6+1)\,/\,\lfloor 10^2/2\rfloor = 27/50 = 54\%~~~~~
  \\[-2em]
\end{eqnarray*}

\vspace{0em}
\subsection{Example~3}\label{ex:3}

Computation of bWER (Eq.\,\ref{eq:werBoW}) for a reference transcript
$X$ and two hypotheses $Y$ and $Z$, and its relation with the naive
bag-of-words WER, $\beta$WER (Eq.\,\ref{eq:betaWER}) and with the
classical WER (Eq.\,\ref{eq:lineWER} or \ref{eq:fullPageWER}).
In both cases, the number of unavoidable insertions is
$b=\big||X|-|Y|\big|=\big||X|-|Z|\big|=14-13=1$.
\vspace{-1.5em}
\begin{eqnarray*}\label{eq:example3}
~~
 X\!&=&\!\text{\sf\small
      to be or not to be that is the question that needs be answered}\\[-0.2em]
 Y\!&=&\!\text{\sf\small
           the question that needs be answered is to be or not to be}\\[-0.2em]
 Z\!&=&\!\text{\sf\small
              to be or   not to be, that is the question to be answered}\\[0.5em]
  \beta\!\WER(X,Y)\!&=&    ~~~~~~~~~~~~~~~~~1/14 =~~7.1\% ~~~~~~~\,
  \beta\!\WER(X,Z)   =     ~~~~~~~~~~~~~~~~~5/14 = 35.7\% \\
        \bWER(X,Y)\!&=&\,(1+1)/(2\!\cdot\!14)    =~~7.1\% ~~~~~~~
        \bWER(X,Z)   =  ~(1+5)/(2\!\cdot\!14)    = 21.4\% \\
         \WER(X,Y)\!&=&\!\!          (0+11+1)/14 = 85.7\% ~~~~~~~~~
         \WER(X,Z)   =               ~(0+2+1)/14 = 21.4\%
\end{eqnarray*}

\vspace{0.3em}
\noindent\emph{Example 3a}.
The bWER can considerably underestimate what might be considered  
``true'' word recognition errors which, in this example, would be
$6/10 = 60\%$:

\vspace{-1.3em}
\begin{eqnarray*}\label{eq:example3b}
~~~~
 X\!\!&=&\!\text{\sf\small
   to be  or not to be, that is the  question}
                      ~~~~~~~~~~~~~~~~~~~~~~~~~~~~~~~\\[-0.2em]
 Y\!\!&=&\!\text{\sf\small
    to be, to not or be  the  is that question
}\\[0em]
 \bWER(X,Y)\!&=& (0+0)/(2\!\cdot\!20) = 0\%
\end{eqnarray*}
\vspace{-1.5em}

\medskip
\subsection{Example~4}\label{ex:4}

Computation of \,hWER for the same texts used in
Example~3~(\ref{ex:3}).  As in Example~2~(\ref{ex:2}), here
$\hat{\mathcal{A}}'(\cdot,\cdot)$ are original alignments obtained as
a byproduct of Eq.\,\eqref{eq:hungCER} and used in
Eq.\,\eqref{eq:hungWER} to compute hWER, and
$\hat{\mathcal{A}}(\cdot,\cdot)$ are the ones used to compute NSFDs
after word renumbering to circumvent the effects of insertion and/or
deletions.
Notice that the values of $\hWER(X,Y)$ and $\hWER(X,Z) $ are identical
to the corresponding bWER values of Example~3.

\begin{figure}[htb]\centering
\!\!\!\!\includegraphics[height=.29\linewidth]{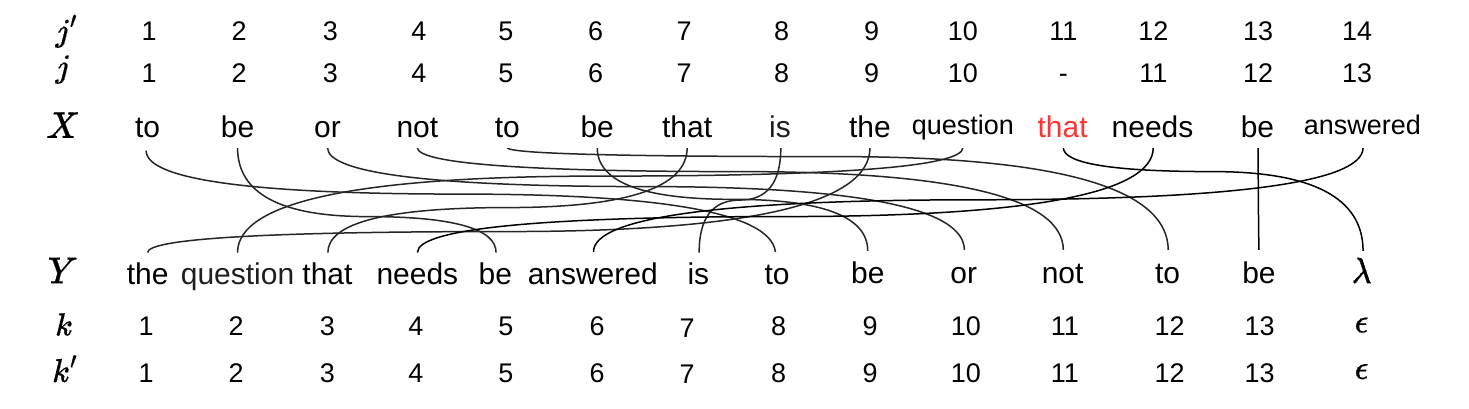}%
\vspace{-1.5em}
\end{figure}

\vspace{-1em}
\begin{eqnarray*}\label{eq:example3c}
  \hat{\mathcal{A}}'(X,Y)\!&=&\!\!
  \text{\small (1,8),(2,5),(3,10),(4,11),(5,12),(6,9),(7,3),(8,7),(9,1),}
  \text{\small (10,2),(11,$\epsilon$),(12,4),(13,13),(14,6)}\\
  \hat{\mathcal{A}}(X,Y)\!&=&\!\!
  \text{\small (1,8),(2,5),(3,10),(4,11),(5,12),(6,9),(7,3),(8,7),(9,1),}
  \text{\small (10,2),(~~-~,$\epsilon$),(11,4),(12,13),(13,6)}\\[0.2em]
   \WER(X,Y)\!\!&=&\!\! (0+11+1)/14  =  85.7\%\\
  \hWER(X,Y)\!\!&=&\!\! (0+~0+\,1)/14 = ~~7.1\%\\
   \rho(X,Y)\!\!&=&\! 71/\lfloor14^2/2\rfloor ~~~~~~\,= 72.4\%~~~~~~~~~~~~~~~~~
\end{eqnarray*}

\vspace{2em}

\begin{figure}[htb]\centering
\!\!\!\!\includegraphics[height=.29\linewidth]{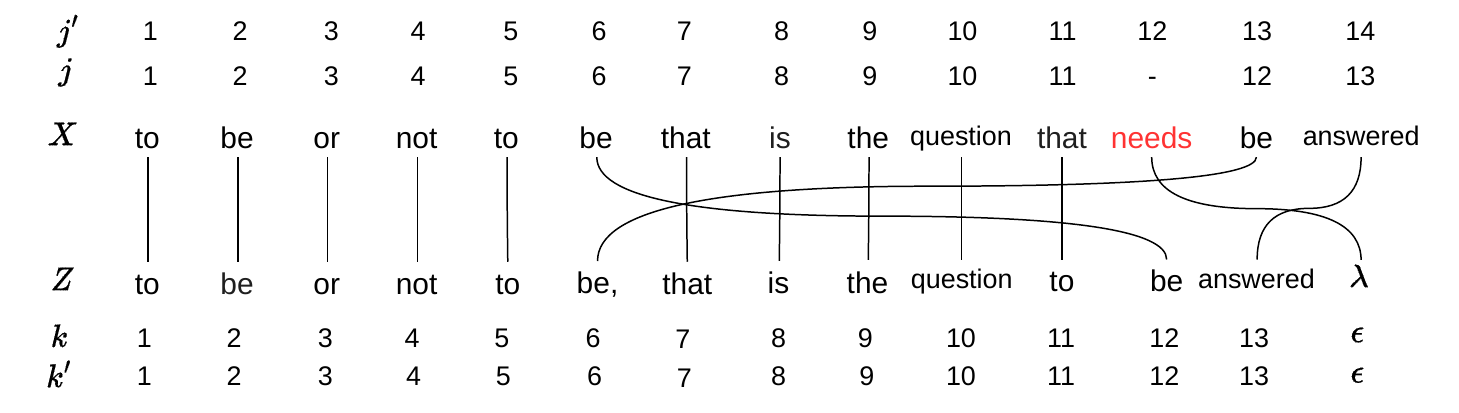}%
\vspace{-3em}
\end{figure}

\begin{eqnarray*}\label{eq:example3c}
  \hat{\mathcal{A}}'(X,Z)\!&=&\!\!
  \text{\small (1,1),(2,2),(3,3),(4,4),(5,5),(6,12),(7,7),(8,8),(9,9),}
  \text{\small (10,10),(11,11),(12,$\epsilon$),(13,6),(14,13)}\\
  \hat{\mathcal{A}}(X,Z)\!&=&\!\!
  \text{\small (1,1),(2,2),(3,3),(4,4),(5,5),(6,12),(7,7),(8,8),(9,9),}
  \text{\small (10,10),(11,11),(~~-~,$\epsilon$),(12,6),(13,13)}\\[0.2em]
   \WER(X,Z)    &=&     (0+2+1)/14 =  21.4\% \\
  \hWER(X,Z)    &=&     (0+2+1)/14 =  21.4\% \\
   \rho(X,Z)    &=&   15/\lfloor14^2/2\rfloor ~~~~~\,= 13.3\%
\end{eqnarray*}

\medskip
\subsection{Impact of multiple word instances and ties in NSFD}\label{ex:5}

When multiple instances of some word exist in $X$ and/or in $Y$, as in
the examples of~\ref{ex:4}, the HA is free to pair any matching
instances, as long as the values of $\mathrm{d_h(X,Y)}$ are the same.
In other words, there may be multiple alignments which provide the
same optimal result for Eq\,\eqref{eq:hungCER} and the HA has no means
to decide which one would be more consistent with the positions of
these words in the RO of the compared texts.

For instance, in Example~4, $\hat{\mathcal{A}}(X,Z)$ pairs
$X_6\!=\,$\textsf{\small``be''} with
$Z_{12}\!=\,$\textsf{\small``be''} and
$X_{13}\!=\,$\textsf{\small``be''} with
$Z_6\!=\,$\textsf{\small``be,''}.
Because of these pairings, the resulting NSFD, $\rho(X,Z)\!=\!13.3\%$,
is exceedingly high, taking into account that $X$ and $Z$ are almost
in the same RO.
Clearly, more consistent or ``natural'' pairings with the same
$\mathrm{d_h(X,Y)}$ are: $X_6\!=\,$\textsf{\small``be''} with
$Z_6\!=\,$\textsf{\small``be,''} and $X_{13}\!=\,$\textsf{\small``be''}
with $Z_{12}\!=\,$\textsf{\small``be''}.
A complete alternative (renumbered) alignment, with identical
$\mathrm{d_h}$ (and $\,\hWER$), would be:
%
\begin{figure}[htb]\centering
\!\!\!\!\includegraphics[height=.29\linewidth]{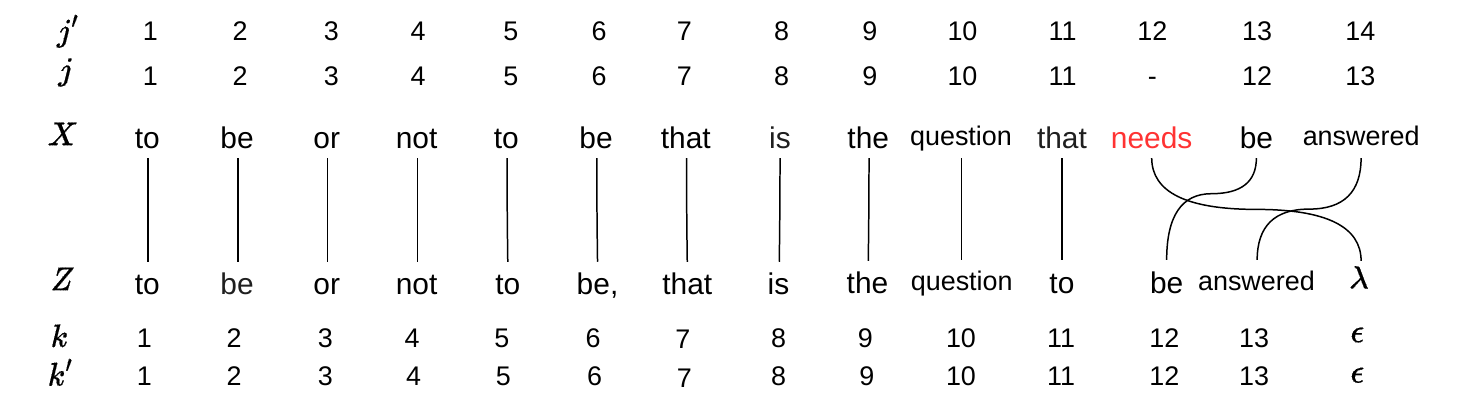}%
\vspace{-0.5em}
\end{figure}

\noindent
$\hat{\mathcal{A}}(X,Z)\!=$
{\small (1,1), (2,2), (3,3), (4,4), (5,5), (6,6), (7,7), (8,8), }%
{\small (9,9), (10,10), (11,11), (~-~,$\epsilon$), (12,12), (13,13)}.\!\\[0.5em]

\vspace{-0.5em}
The NSFD of such an alignment is much lower:
$\rho(X,Z)=1/\lfloor14^2/2\rfloor = 1.0\%$, which better reflects the
very minor RO discrepancy between $X$ and $Z$.

\section{Software Tools and Datasets}\label{ap:software}

The software, with the implementation of the metrics employed
to evaluate End-to-End HTR approaches, is freely available to download
and use for replicating the results reported in this paper.%
\footnote{\url{https://github.com/PRHLT/E2EHTREval.git}}

Most of its functionalities have been programmed in python, like
computation of the NSFD metric and the building of the
edit-distance-based cost-matrix with the proposed regularisation
factor of Eq.\,\eqref{eq:hungCER} for using with the HA. Regarding the
time-critical HA computation, we employ the implementation provided by
the \emph{scipy library}%
\footnote{\url{https://docs.scipy.org/doc/scipy/reference/generated/scipy.optimize.linear_sum_assignment.html}}
implemented in C and with a python-wrapper, which is based on the one
described in\,\cite{crouse2016implementing}.  For the also
time-critical Levenshtein edit-distance computation, it was employed
an extended version of \texttt{fasterwer}%
\footnote{\url{https://github.com/PRHLT/fastwer}} 
(forked from the original one%
\footnote{\url{https://github.com/kahne/fastwer}}), a library written
in C++ and wrapped in python for ease of use. In this library we have
also included support for UTF-8 encoding as well as others
time-critical functionalities like the implementation of bag-of-words
(see Eq.\,\eqref{eq:werBoW}) based on hashing for faster computation,
and the implementation of the backtrace algorithm to obtain the
aligning-path through a minimum edit-distance between reference and
hypothesis strings.

The datasets used throughout this work can be downloaded most of them
from the \textsc{zenodo} platforms:
ICFHR14%
\footnote{\url{https://zenodo.org/record/44519}},
IAMDB%
\footnote{\url{https://fki.tic.heia-fr.ch/databases/iam-handwriting-database}},
ICFHR16%
\footnote{\url{http://doi.org/10.5281/zenodo.1164045}},
ICDAR17%
\footnote{\url{http://doi.org/10.5281/zenodo.835489}} and
FCR%
\footnote{\url{https://zenodo.org/record/3945088\#.Y3u_tkjMLZ8}}.

\end{document}